\documentclass[sigconf,nonacm]{acmart}
\usepackage{booktabs} 
\usepackage{algorithm}
\usepackage{multirow}
\usepackage{booktabs}
\usepackage{subcaption}
\usepackage{algpseudocode}
\AtBeginDocument{%
  }

\acmConference[Conference acronym 'XX]{Make sure to enter the correct
  conference title from your rights confirmation email}{June 03--05,
  2018}{Woodstock, NY}
\acmISBN{978-1-4503-XXXX-X/2018/06}

\begin{document}

\title{Privacy Auditing Synthetic Data Release through Local Likelihood Attacks}

\author{Joshua Ward}
\affiliation{%
  \institution{University of California, Los Angeles}
  \city{Los Angeles}
  \state{CA}
  \country{USA}
}
 
\author{Chi-Hua Wang}
\affiliation{%
  \institution{Purdue University}
  \city{West Lafayette}
  \state{IN}
  \country{USA}
}
 
\author{Guang Cheng}
\affiliation{%
  \institution{University of California, Los Angeles}
  \city{Los Angeles}
  \state{CA}
  \country{USA}
}

\renewcommand{\shortauthors}{Ward et al.}

\begin{abstract}
Auditing the privacy leakage of synthetic data is an important but unresolved problem. Existing privacy auditing frameworks for synthetic data rely on heuristics and unrealistic assumptions about model access, offering limited ability to describe or detect the privacy exposure of training data through synthetic data release. In this paper, we study designing membership inference attacks (MIAs) that specifically exploit the observation that tabular generative models tend to significantly overfit to certain regions of the training distribution. 

We propose \emph{Generative Likelihood Ratio Attack} (Gen-LRA), a novel, computationally efficient No-Box MIA that, with no assumption of model knowledge or access, formulates its attack by evaluating the influence a test observation has on a surrogate model's estimate of a local likelihood ratio over the synthetic data. We develop a theoretical framework for the attack: we show that the Gen-LRA score admits a closed-form characterization as a localized density-ratio statistic, and we prove that under a general model of local overfitting it produces a provable mean-score gap between members and non-members, yielding testable predictions for when the attack should succeed. We validate these predictions in a controlled simulation study and assess Gen-LRA against a comprehensive benchmark spanning diverse datasets, generative model architectures, and attack parameters. Across metrics, Gen-LRA consistently dominates competing MIAs, with especially strong gains at low false positive rates. These results underscore Gen-LRA's effectiveness as a privacy auditing tool for the release of synthetic data, and highlight the significant privacy risks posed by generative model overfitting in real-world applications.
\end{abstract}



\maketitle

\section{Introduction}

Real-world tabular data is often privacy-sensitive to the individual observations that compose these samples, hindering their ability to be shared in open-science efforts that can aid in new research and improve reproducibility. A promise of generative modeling is that models trained on sensitive data can produce samples that preserve the privacy of the training set while maintaining much of its intrinsic statistical information, enabling responsible release to a third party. In practice, a wide array of methodologies have been proposed to accomplish synthetic data release involving modifying loss functions \citep{Abadi_2016, WANG2022306}, creating new generative model architectures \citep{yoon2018pategan,PMID:32167919}, and studying data release strategies \citep{10.5555/2095116.2095131,10.1007/978-3-642-28914-9_19,9458927} to provide differential privacy guarantee. In another direction, a variety of methods have been proposed that maximize the fidelity of synthetic data and argue that privacy is satisfied through ad-hoc similarity metrics \citep{zhao2021ctab,guillaudeux2023patient, liu2023tabular, solatorio2023realtabformer}.

To audit the empirical privacy of synthetic data generators, membership inference attacks (MIAs) have recently been extended from traditional machine learning models to synthetic tabular data. Here, privacy auditing is framed as an adversarial game: given specific constraints defined by a threat model, an attacker attempts to determine whether a test observation belongs to a model's training dataset exploiting some notion of model failure \citep{Shokri,ganleaks,Carlini2021MembershipIA}. A successful attack represents a concrete privacy breach with clear real-world implications, where other similarity-based metrics have been shown to fail to capture privacy risk \citep{platzer2021holdout, ganev2023inadequacy, ward2024dataplagiarismindexcharacterizing}.

While promising, MIAs for generative models and synthetic data release have seen limited success. Previous work has often relied on distance- or density-based heuristics, or has made additional assumptions about model query access that are unrealistic to the release setting and do not scale computationally to modern architectures. Moreover, most existing attacks are proposed without a theoretical account of what they measure or when they should succeed, making it difficult to assess whether empirical performance reflects a principled design or a coincidence of benchmark choice. We address both gaps in this work.

We focus on studying membership inference for the release of synthetic data in a No-Box Threat Model \citep{houssiau2022tapas}. In this setting, we make no adversarial assumptions of knowledge about model architecture, access, or training parameters, mimicking real-world scenarios of parties following best practices for releasing synthetic data in domains like healthcare and finance. Under this threat model, we derive a powerful MIA called \emph{Generative Likelihood Ratio Attack} (Gen-LRA), which constructs an influence function formulated from likelihood ratio estimation to target privacy leakage that occurs through model overfitting. We develop a theoretical framework that characterizes what Gen-LRA computes and predicts when it succeeds, validate these predictions in a controlled simulation study, and demonstrate empirical dominance across a large-scale benchmark of 1{,}525 synthetic dataset configurations.

\textbf{Contributions}:
\begin{enumerate}
    \item \emph{A novel attack.} We introduce Gen-LRA, a No-Box MIA that formulates membership inference as the influence of a test point on a surrogate model's estimate of the likelihood ratio over a local subset of synthetic data. To our knowledge, this is the first MIA for tabular synthetic data to explicitly adapt an influence-function framework to the No-Box setting.

    \item \emph{Theoretical framework.} We develop three theoretical results supporting Gen-LRA: (i) an invariance property showing that the log-likelihood-ratio score is insensitive to invertible reparametrizations of the data; (ii) a closed-form characterization showing that Gen-LRA is, to leading order, a localized density-ratio statistic, making explicit what the attack measures and why it differs from prior density-based MIAs; and (iii) a mean-score-gap result that, under a general model of local overfitting, produces a provable separation between members and non-members and identifies the structural quantities that govern the attack's success.
    
    \item \emph{Simulation validation.} We employ a controlled simulation that directly instantiates the overfitting model, validating our theoretical results and characterizing when membership inference is and is not feasible. We find Gen-LRA is uniquely tunable across overfitting regimes, and the attack outperforms existing methods particularly when a generator approximately memorizes its training data.    
    
    \item \emph{Empirical benchmark.} We show that Gen-LRA broadly outperforms competing MIAs across a benchmark of 35 datasets, 9 generative architectures, and 9 attacks, comprising 1{,}525 unique synthetic dataset configurations and over 10{,}000 attack runs. Gen-LRA is the top-ranked attack on more than $2\times$ as many runs as any baseline across both AUC and TPR at all fixed false positive rates  (Table~\ref{tab:rank}), and achieves the highest mean values across the top 100 highest-leakage runs on every metric (Table~\ref{tab:mean}). We additionally perform ablations across encoding strategies, surrogate density estimators, and locality and bandwidth choices, finding that Gen-LRA is robust across these design decisions.
\end{enumerate}
\section{Membership Inference Attacks Preliminaries}
\label{sec:formalism}
 
In this work, we study the Membership Inference Attack Game in the context of synthetic data release. The objective of this game is to determine whether a particular data point was included in the original training dataset by examining the outputs of a generative model. We first introduce the formal definition of the Membership Inference Attack Game modeled after \cite{Shokri}.
\subsection{Membership Inference Attack Game}

\paragraph{Definition (Membership Inference Attack Game)}
The game proceeds between a challenger $\mathcal{C}$ and an adversary $\mathcal{A}$ as follows:
\begin{enumerate}
    \item The challenger samples a training dataset $T = \{x_i\}_{i=1}^n$ from the population distribution $x_i \sim \mathbb{P}$ over the domain $\mathcal{X} = \mathcal{X}_1 \times \cdots \times \mathcal{X}_d$, where each attribute domain $\mathcal{X}_j$ is either numeric (i.e., $\mathcal{X}_j \subseteq \mathbb{R}$) or categorical (i.e., $\mathcal{X}_j$ is a finite discrete set), and uses $T$ to train a tabular generative model $G \leftarrow \mathcal{T}(T)$. The generative model $G$ produces synthetic dataset $S$.
    \item The challenger flips a bit $b \in \{0, 1\}$. If $b = 0$, the challenger samples a test observation $x^\star \in \mathcal{X}$ from the population distribution $\mathbb{P}$. Otherwise, the challenger selects the test observation $x^\star$ from the training set $T$.
    \item The challenger sends the test observation $x^\star$ to the adversary $\mathcal{A}$.
    \item The adversary has access to some information defined by a threat model and uses this information to output a guess $\hat{b} \leftarrow \mathcal{A}(x^\star)$.
    \item The output of the game is $1$ if $\hat{b} = b$, and $0$ otherwise. The adversary wins if $\hat{b} = b$, i.e., if it correctly identifies whether $x^\star$ was part of the training set $T$ or a sampled point from $\mathbb{P}$.
\end{enumerate}
 
\paragraph{Adversary's goal and capabilities.}
The adversary $\mathcal{A}$ aims to determine whether a specific data point $x^\star$ was part of the original training dataset $T$ or was drawn from the population distribution $\mathbb{P}$. The adversary can use any available information to construct a scoring function that classifies the membership of $x^\star$ defined by a threat model. The performance of this classifier, evaluated with binary classification metrics, measures the privacy leakage of the training data for $G$ through $S$. Formally, a membership inference attack can be expressed as
\begin{equation}
\mathcal{A}(x^\star) \;=\; \mathbb{I}\!\left[f(x^\star) > \gamma\right],
\label{eq:membership_prediction}
\end{equation}
where $\mathbb{I}$ is the indicator function, $f(x^\star)$ is a scoring function of $x^\star$, and $\gamma$ is an adjustable decision threshold.
 \subsection{Threat Model}

In this work, we consider a ``No-Box''~\citep{houssiau2022tapas} threat model in which the adversary has no access to the internal structure, parameters, or sampling mechanism of the generative model. The attack must instead be constructed using only two observed datasets: the released synthetic dataset $S \sim G(T)$, and an independently collected reference dataset $R \sim \mathbb{P}$ drawn from the same underlying population. The adversary is not granted access to knowledge of the model implementation nor can they issue queries to a trained generator. This reflects deployment scenarios in which organizations release synthetic data for downstream analysis while keeping all model knowledge confidential. The synthetic dataset $S$ serves as the only potential leakage surface, and the reference set $R$ provides a statistical anchor for the population. A reference dataset is commonly assumed in No-Box attacks on synthetic data~\citep{ganleaks, houssiau2022tapas, vanbreugel2023membership, ward2024dataplagiarismindexcharacterizing} as well as in MIAs for supervised learning models~\citep{Carlini2021MembershipIA, ye2022, Sablayrolles2019WhiteboxVB}, and represents a scenario in which an adversary may be able to find comparable data in the real world---for example via open-source datasets, paid collection, or domain knowledge.

Three considerations motivate our focus on this threat model. First, it is the most realistic setting for synthetic data release. Shadow-box attacks \cite{groundhog, houssiau2022tapas,Meeus_2024} rely on the adversary knowing the generator's architecture or implementation in order to train surrogate models, but a defender following release best practices can trivially neutralize such attacks by simply withholding this information~\citep{golob2024privacyvulnerabilitiesmarginalsbasedsynthetic}. The estimated privacy leakage of these attacks is therefore not conditioned on the knowledge an adversary would realistically have. Second, shadow-box and white-box approaches do not scale computationally to modern tabular generators. Diffusion-based and transformer-based architectures can take hours to train for a single model, and attacks that require fitting hundreds of surrogate models per auditing run become infeasible as the generator's training cost grows. Third, No-Box attacks study leakage through the synthetic data itself rather than through any particular model, making them model-agnostic. This property makes them useful both as a practical privacy-auditing tool that a third party can run on released synthetic data, and as a research tool for understanding the privacy behavior of generative models independent of their specific architectures.

\section{Related Works}
\label{sec:related}
 
\subsection{Assessing Overfitting in Tabular Generative Models}
\label{sec:related_overfit}
 
Several measures have been developed to assess the fitness of tabular synthetic data, particularly from a privacy perspective. These metrics generally aim to measure the similarity between the training and synthetic datasets, with the ideal outcome being that the synthetic data is neither too similar to the training data nor too different. A widely used metric for this purpose is the Distance to Closest Record\footnote{DCR in the similarity-metric case compares a training point to a synthetic point. \citet{ganleaks} propose an MIA in which the scoring function is a distance computation for a test point and a synthetic point; in all other sections of the paper we use DCR to refer to the MIA.}~\citep{park2018data,liu2023tabular, lu2019empirical, yale2019assessing, zhao2021ctab, guillaudeux2023patient}, which measures the distance from each training point to its nearest neighbor in the synthetic dataset and averages across training points. Another commonly used metric is the Identical Matching Score (IMS)~\citep{lu2019empirical, ims1, ims2}, which measures the proportion of identical records between the training and synthetic datasets. While these measures can be useful for describing overfitness from a distribution-level quality or model-generalization perspective, they do not characterize privacy risk: there is no assumed threat model and they are not evaluated against non-member examples.
 
Exploiting overfitting as a source of privacy leakage has been documented in specific generators. \citet{vanbreugel2023membership} showed that TVAE~\citep{Xu2019ModelingTD} overfits to minority-class examples in a medical training dataset, leaking their privacy. \citet{ward2024dataplagiarismindexcharacterizing} found that Tab-DDPM~\citep{tabddpm} heavily replicates training records from certain demographic subgroups when generating synthetic data for the Adult dataset. These findings motivate our approach of designing an attack that explicitly targets the local overfitting behavior of tabular generative models (see Section~\ref{sec:genlra}).
 
\subsection{MIAs for Tabular Generative Models}
\label{sec:related_mias}
 
Membership inference attacks explicitly characterize the empirical privacy risk of a machine learning model~\citep{Song2020SystematicEO, Yeom}. MIAs were originally developed for attacking supervised learning classifiers~\citep{Shokri}, where the general idea is to query a model with different observations to learn patterns in its class-probability outputs. Membership is then inferred by comparing the outputs of the model to outputs from reference models in various ways~\cite{Carlini2021MembershipIA, Long, Sablayrolles2019WhiteboxVB, watson2022on, ye2022, zarifzadeh2024}.
 
To adapt to the structural differences of generative models, a range of MIAs for tabular generators have been proposed that employ different threat models and strategies~\citep{ganleaks, Hayes2017LOGANMI, Hilprecht2019MonteCA, houssiau2022tapas, Meeus_2024, groundhog, vanbreugel2023membership, ward2024dataplagiarismindexcharacterizing}. Gen-LRA is most closely related to DOMIAS~\citep{vanbreugel2023membership} and to a line of work that extends query-based attacks to tabular generative models~\cite{houssiau2022tapas, Meeus_2024, groundhog}.
 
\paragraph{Relation to DOMIAS}

DOMIAS follows the same threat-model assumptions as Gen-LRA and similarly defines its scoring function in \eqref{eq:membership_prediction} as a density ratio, $\frac{p_S(x^*)}{p_R(x^*)}$. However, its theoretical foundation is limited because it effectively tests the wrong membership hypothesis: rather than testing whether the inclusion of $x^*$ in the reference data influences the released synthetic dataset, it tests only whether $x^*$ lies in a region where the synthetic distribution is denser than the reference distribution. In contrast, Gen-LRA directly targets the membership inference problem by measuring the effect of specifically including $x^*$ on $R$, angling with the true hypothesis of interest. Moreover, while DOMIAS relies on a single-point density estimate, Gen-LRA aggregates evidence over a local region, allowing it to capture diffuse memorization signals and incorporate substantially more information. Our theoretical analysis, simulations, and empirical benchmarks all support this distinction: because DOMIAS tests the wrong hypothesis, its performance degrades substantially in a variety of overfitting regimes, whereas Gen-LRA remains robust and consistently outperforms it.
 
\paragraph{Relation to query-based attacks}
\cite{groundhog}, \citet{houssiau2022tapas}, and \citet{Meeus_2024} propose query-based attacks on tabular generators that additionally assume the adversary has knowledge of the \emph{implementation} of the target model. In these methods, an attacker trains many versions of the model on $R \cup \{x^\star\}$ and $R$, generates many synthetic datasets, represents each synthetic dataset by summary statistics or histograms, and trains a classifier to distinguish between the two regimes. 

Gen-LRA improves on these attacks in two ways. First, they are unsuitable for auditing privacy in a release setting because they are trivially defeated if the defender chooses not to disclose the architecture; indeed, \citet{golob2024privacyvulnerabilitiesmarginalsbasedsynthetic} has shown significant privacy leakage arising from architectural disclosure, leading to the recommendation that data-releasing parties disclose as little model information as possible. Gen-LRA makes no assumption about model implementation. Second, query-based attacks are computationally expensive: they require training $(N_{\text{test}} + 1) \times N_{\text{surrogate}}$ separate models to audit a single generator, which is impractical as large diffusion and language model architectures become more common. Gen-LRA requires only $N_{\text{test}} + 1$ density estimators to be fit, which is substantially cheaper.

\section{Generative Likelihood Ratio Attack}
\label{sec:genlra}
 
In this section, we propose \emph{Generative Likelihood Ratio Attack} (Gen-LRA), a membership inference attack designed to detect membership leakage in synthetic data through a statistical notion of \emph{likelihood influence}. Unlike prior MIAs that evaluate the density or distance of a test point itself, Gen-LRA frames membership inference as a measurement of how much the test point $x^\star$ influences an estimate of the likelihood of $S$. The central idea is that if $x^\star$ belonged to the training set and the generative model is overfit, then adding $x^\star$ to a surrogate density estimator should increase the estimated likelihood of the synthetic dataset.
 
We develop Gen-LRA in six steps. Section~\ref{sec:infl} reviews empirical influence functions, which provide the theoretical foundation for the attack. Section~\ref{sec:attack_surface} defines the Gen-LRA score as a log-likelihood-ratio influence function. Section~\ref{sec:why_lr} justifies the log-likelihood-ratio form via Neyman-Pearson optimality and an invariance property. Sections~\ref{sec:what_measures} and \ref{sec:when_works} establish the two main theoretical results of the paper: a closed-form characterization of the Gen-LRA score, and a mean-score-gap result quantifying the attack's power under a general overfitting model. Section~\ref{sec:implementation} describes the practical implementation, with each design choice motivated by the preceding theory.
 
\subsection{Empirical Influence Functions}
\label{sec:infl}
 
Influence functions, originally developed in the field of Robust Statistics~\citep{hampel1974influence, cook1986residuals}, measure how statistical estimates change when the underlying data distribution is perturbed. The influence function for an estimator $\theta$ applied to a distribution $F$ is
\begin{equation}
\mathcal{I}(x^\star, F, \theta) \;=\; \lim_{\epsilon \to 0} \frac{\theta((1-\epsilon)F + \epsilon\delta_{x^\star}) - \theta(F)}{\epsilon},
\end{equation}
where $\delta_{x^\star}$ is the Dirac measure placing unit mass at $x^\star$. This quantity captures the sensitivity of $\theta$ to infinitesimal perturbations of $F$ at $x^\star$.
 
In the empirical setting with finite samples $\mathcal{D} = \{x_1, \ldots, x_n\}$, the empirical influence function is defined by evaluating how an estimate changes when a point is added:
\begin{equation}
\hat{\mathcal{I}}(x^\star, \mathcal{D}, \theta) \;=\; \theta(\mathcal{D} \cup \{x^\star\}) - \theta(\mathcal{D}).
\end{equation}
For supervised learning models, this difference is typically measured on loss or empirical risk~\citep{pmlr-v70-koh17a}. In an MIA for tabular generative models with no model access, however, measures of loss are not available. Rather than examining how $x^\star$ affects model parameters, Gen-LRA considers the influence of $x^\star$ on the likelihood assigned to generated samples $S$ by a surrogate estimator.
 
\subsection{Likelihood Influence as an Attack Surface}
\label{sec:attack_surface}
 
\begin{figure*}
    \centering
    \includegraphics[width=.8\textwidth]{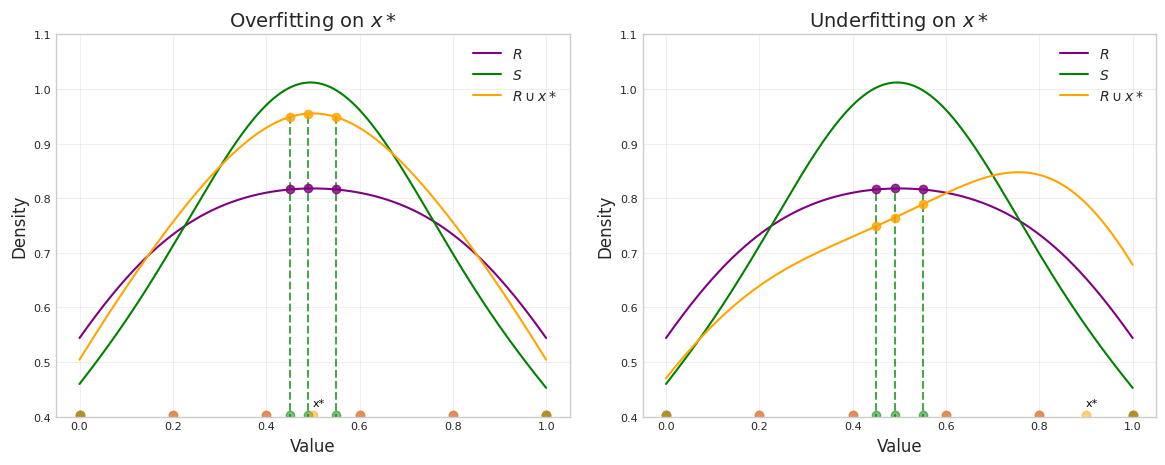}
    \caption{A geometric intuition for Gen-LRA with a 1-dimensional toy example. We visualize the KDE plots of $R, R \cup \{x^\star\}, S$ as well as the estimated densities of the synthetic observations over $R$ and $R \cup \{x^\star\}$. Left: $x^\star = 0.5$. The likelihood of the synthetic observations (product of orange intersections) is higher under the density estimate of $R \cup \{x^\star\}$ than $R$ (product of purple intersections), so the attack concludes $x^\star \in T$. Right: $x^\star = 0.9$; the opposite holds, so $x^\star \notin T$ is concluded.}
    \label{fig:geometric_intuition}
        \Description{Two side-by-side density plots illustrating Gen-LRA on a 1D example. Each panel shows three overlapping KDE curves for the reference set R, the synthetic set S, and the augmented set R union x-star, with vertical dashed lines marking where each synthetic point intersects the R and R union x-star densities. In the left panel, x-star equals 0.5 and sits at the peak of S; the augmented-density intersections are higher than the reference-density intersections, so the attack concludes x-star is a member. In the right panel, x-star equals 0.9 and sits in the tail of S; the augmented-density intersections are lower, so the attack concludes x-star is not a member.}
\end{figure*}

Recalling that $T, R \sim \mathbb{P}$ and that our goal is to infer whether $x^\star \in T$ given $S$ and $R$, we hypothesize that if the generator is overfit \footnote{We use \emph{overfitting} for the generator's behavior of placing excess mass near training points relative to the true population, and \emph{memorization} for its geometric manifestation in the synthetic data. Both notions are formalized in Section~\ref{sec:when_works}.} then $S$ carries signal of $x^\star$ specifically when $x^\star \in T$. Gen-LRA measures this overfitness by formalizing an influence function on the estimated log-likelihood of $S$ under two surrogate models: one fit on the reference dataset $R$, and another fit on the augmented dataset $R \cup \{x^\star\}$:
\begin{equation}
\hat{\mathcal{I}}(x^\star; R, S) \;:=\; \log \hat{p}(S \mid R \cup \{x^\star\}) - \log \hat{p}(S \mid R).
\label{eq:abstract_loglik_diff}
\end{equation}
Intuitively (see Figure~\ref{fig:geometric_intuition}), if adding $x^\star$ to $R$ substantially increases the estimated likelihood of $S$, this suggests $x^\star$ contributed to the generative process. If the likelihood is unchanged or decreases, it implies $x^\star \notin T$. Using $\hat{\mathcal{I}}(x^\star; R, S)$ as the scoring function $f$ in the membership prediction rule~\eqref{eq:membership_prediction} defines the Gen-LRA attack.
 
\subsection{Justification of the Log-Likelihood Ratio}
\label{sec:why_lr}
 
The influence function in~\eqref{eq:abstract_loglik_diff} is one of many possible functionals of $\hat{p}(S \mid R \cup \{x^\star\})$ and $\hat{p}(S \mid R)$. Two properties, one statistical and one geometric, justify the choice of the log-likelihood ratio.
 
\paragraph{Neyman-Pearson optimality}
The membership inference game is fundamentally a binary hypothesis test: $H_0: x^\star \sim P$ against $H_1: x^\star \in T$. By the Neyman-Pearson lemma, the likelihood ratio test is uniformly most powerful among all tests at a given false positive rate. This principle has long been central to the MIA literature: \citet{Shokri} frame membership inference as a classification problem whose Bayes-optimal solution reduces to a likelihood ratio, \citet{Sablayrolles2019WhiteboxVB} derive the Bayes-optimal MIA strategy and show it takes a likelihood-ratio form, and \citet{ye2022} argue for calibrated likelihood-ratio statistics as a general design principle.
 
\citet{Carlini2021MembershipIA} make this argument operational, demonstrating that MIAs grounded explicitly in likelihood ratios (rather than ad-hoc distance, confidence, or loss statistics) achieve substantially stronger guarantees in the supervised-learning setting, particularly at the low false positive rates that matter most for practical privacy auditing. Subsequent work has extended this principle to enhanced and low-cost variants~\citep{ye2022, zarifzadeh2024}. Gen-LRA adapts this design principle to the No-Box tabular synthetic data setting: the scoring function in~\eqref{eq:abstract_loglik_diff} is a log-likelihood ratio over $S$ under two hypotheses about the data distribution that generated it, making it the natural Neyman-Pearson analogue for the attack surface available to a No-Box adversary.
 
\paragraph{Invariance to reparametrization}
A second advantage of the log-likelihood-ratio form is that it is invariant to invertible transformations of the data.
 
\begin{theorem}[Invariance of the population influence function]
\label{thm:invariance}
Let $S$ and $R$ be sets of samples and $x^\star$ a test point, with probability distributions on $\mathcal{X} \subseteq \mathbb{R}^d$. For any continuously differentiable invertible function $g: \mathcal{X} \to \mathcal{X}$ with non-vanishing Jacobian, the population-level log-likelihood-ratio influence function is invariant:
\begin{equation}
\mathcal{I}(g(x^\star), g(R), g(S)) \;=\; \mathcal{I}(x^\star, R, S).
\end{equation}
\end{theorem}
 
The proof (Appendix~\ref{app:invariance_proof}) follows from the change-of-variables formula and cancellation of Jacobian terms. This invariance is practically important: tabular data is typically preprocessed through one-hot encoding, scaling, or learned embeddings, and an attack whose score depends on these arbitrary choices would be unreliable. Competing attacks for example based on distance heuristics such as Distance to Closest Record do not satisfy this property.
 
\subsection{Closed-Form Characterization of the Attack Score}
\label{sec:what_measures}
 
Having justified the log-likelihood-ratio form, we now analyze what Gen-LRA computes. To make this analysis concrete, we specialize to a particular choice of surrogate density estimator and localization strategy. Throughout the remainder of this section, we take $\hat{p}_R$ to be a Gaussian Kernel Density Estimator (KDE) fit on the reference set $R$, and we localize the attack statistic to the $k$-nearest synthetic neighbors of $x^\star$ rather than summing over all of $S$.
 
Both of these choices are justified empirically and discussed in full in Section~\ref{sec:implementation}; briefly, Gaussian KDEs admit a closed-form expansion that makes the subsequent analysis tractable and empirically outperform more expressive density estimators (Section~\ref{subsec:deeplearningdensityestimation}), while localization concentrates the statistic on the region where overfitting signal can exist.
 
Formally, let $\hat{p}_R$ denote a Gaussian KDE fit on reference set $R$ with $|R| = m$ and bandwidth $h$, let $K_h(u) = h^{-d} K(u/h)$ be the Gaussian kernel in $\mathbb{R}^d$, and let $S_k(x^\star) \subseteq S$ be the set of $k$ synthetic points in $S$ nearest to $x^\star$. The Gen-LRA score is then
\begin{equation}
\hat{\mathcal{I}}(x^\star; R, S) \;=\; \sum_{s \in S_k(x^\star)} \left[\log \hat{p}_{R \cup \{x^\star\}}(s) - \log \hat{p}_R(s)\right].
\end{equation}
Our first theoretical result characterizes this quantity in closed form.
 
\begin{theorem}[Closed-form approximation]
\label{thm:closedform}
Suppose $\hat{p}_R(s) \geq c > 0$ for all $s \in S_k(x^\star)$. Then, for any fixed $k$ or any $k = o(m)$, the Gen-LRA score satisfies
\begin{equation}
\hat{\mathcal{I}}(x^\star; R, S) \;=\; \frac{1}{m+1} \sum_{s \in S_k(x^\star)} \left[\frac{K_h(s - x^\star)}{\hat{p}_R(s)} - 1\right] \;+\; O_p(m^{-2}).
\label{eq:closedform}
\end{equation}
\end{theorem}
 
The proof (Appendix~\ref{app:thm1_proof}) follows from an exact identity for the augmented KDE together with a Taylor expansion of the logarithm. Theorem~\ref{thm:closedform} shows that Gen-LRA is, to leading order, a sum of localized density-ratio-like quantities. The dominant term $K_h(s - x^\star) / \hat{p}_R(s)$ is large precisely when a synthetic point $s$ is close to $x^\star$ (numerator) \emph{and} lies in a region of low reference density (denominator). This is the fingerprint of memorization: training-induced synthetic points that cluster near $x^\star$ in regions the population would not naturally populate.
 
\paragraph{Comparison to DOMIAS}
DOMIAS~\citep{vanbreugel2023membership} uses the scoring function $\hat{p}_S(x^\star)/\hat{p}_R(x^\star)$---a single density ratio evaluated at $x^\star$. Equation~\ref{eq:closedform} shows that Gen-LRA evaluates an analogous ratio at $k$ synthetic points correlated with memorization of $x^\star$, rather than at $x^\star$ itself. Gen-LRA thus aggregates evidence over a region where the memorization signal is expected to concentrate, while DOMIAS uses only a single-point estimate. This provides a formal explanation for why Gen-LRA and DOMIAS differ despite following the same threat model.
 
\paragraph{Role of localization}
The closed form in~\eqref{eq:closedform} makes clear why localization is essential. The summand is appreciable only at synthetic points $s$ near $x^\star$; points far from $x^\star$ contribute $K_h(s - x^\star) \approx 0$ and thus only the $-1$ baseline term, which adds noise without signal. Restricting the sum to the $k$-nearest synthetic neighbors of $x^\star$ discards these noise-only terms and concentrates the statistic on the region where signal can exist. The parameter $k$ thus serves as a bias-variance trade-off: larger $k$ reduces variance by averaging more terms, while smaller $k$ reduces bias by excluding neighbors outside the memorization region.
 
\subsection{Power Under a Local Overfitting Model}
\label{sec:when_works}
 
Theorem~\ref{thm:closedform} characterizes what Gen-LRA computes; it does not by itself guarantee that the computation separates members from non-members. We now establish that under a general model of local overfitting, the expected Gen-LRA score is provably larger for members than for non-members, and we characterize the conditions under which the gap is largest.
 
\begin{definition}[Local overfitting]
\label{def:localoverfit}
A generator $G$ trained on $T$ exhibits \emph{local overfitting at scale $\rho$ with strength $\epsilon$} if its induced density $p_G$ decomposes as
\begin{equation}
p_G(s) \;=\; p(s) \;+\; \frac{\epsilon}{|T|}\sum_{x_i \in T} \phi_\rho(s; x_i) \;+\; r(s),
\label{eq:overfitmodel}
\end{equation}
where $p$ is the population distribution, $|T|$ denotes the cardinality of the training set, $\phi_\rho(\cdot\,; x_i)$ is a non-negative \emph{excess mass function} concentrated within distance $\rho$ of $x_i$ with $\int \phi_\rho(s; x_i)\, ds = 1$, $\epsilon \in [0,1]$ is the \emph{overfitting strength}, and $r(s)$ is a residual satisfying $\int r(s)\, ds = -\epsilon$.
\end{definition}
 
The formulation is deliberately general: $\phi_\rho$ can be any concentrated non-negative function, a sharp spike for exact replication, a diffuse bump for noisy memorization, or an asymmetric shape reflecting the generator's inductive bias, and $r$ absorbs any displacement of mass from elsewhere in the distribution. The decomposition does not require a specific generator or model; it is purely a description of its output density. 
 
\begin{theorem}[Mean score gap]
\label{thm:meangap}
Let $\mu_1 = \mathbb{E}[\hat{\mathcal{I}}(x^\star; R, S) \mid x^\star \in T]$ and $\mu_0 = \mathbb{E}[\hat{\mathcal{I}}(x^\star; R, S) \mid x^\star \sim P]$ denote the expected Gen-LRA scores for members and non-members, respectively. Under Definition~\ref{def:localoverfit} and the regularity conditions of Theorem~\ref{thm:closedform},
\begin{equation}
\mu_1 - \mu_0 \;\gtrsim\; \frac{k \, \epsilon}{(m+1)\,|T|} \cdot \Psi(h, \rho, x^\star),
\label{eq:meangap}
\end{equation}
where the \emph{signal term} is
\begin{equation}
\Psi(h, \rho, x^\star) \;=\; \mathbb{E}_s\!\left[\frac{K_h(s - x^\star)\,\phi_\rho(s; x^\star)}{\hat{p}_R(s)}\right],
\label{eq:signalterm}
\end{equation}
and $\Psi(h, \rho, x^\star) > 0$ whenever the support of $\phi_\rho(\cdot; x^\star)$ overlaps with the kernel $K_h(\cdot - x^\star)$.
\end{theorem}
 
We defer the proof to Appendix~\ref{app:thm2_proof}. The argument proceeds by substituting the overfitting decomposition~\eqref{eq:overfitmodel} into the leading-order expression from Theorem~\ref{thm:closedform}, and showing that the contributions from the population density $p$ and the residual $r$ cancel in the difference $\mu_1 - \mu_0$, leaving the self-overlap $\Psi$ as the dominant signal.
 
Theorem~\ref{thm:meangap} states that Gen-LRA has nontrivial power against any generator exhibiting local overfitting with $\epsilon > 0$. The size of the mean gap depends on three interpretable quantities, each governing a distinct aspect of the attack's behavior:
\begin{enumerate}
    \item \emph{Linear dependence on the overfitting strength $\epsilon$.} The mean gap of Gen-LRA scales linearly with $\epsilon$: generators that overfit more aggressively are more vulnerable to Gen-LRA, and generators that do not, for example those trained with strong differential privacy guarantees, to which $\phi_\rho$ must be nearly flat, should yield low-quality attacks.
    
    \item \emph{Localization to the memorization region.} The score in~\eqref{eq:closedform} approximates $\Psi$ by summing the integrand over the $k$ nearest synthetic neighbors of $x^\star$, so $k$ controls how many synthetic points are included in the sum. When $k$ is smaller than the number of synthetic points the generator places within distance $\sim \rho$ of $x^\star$, the sum ignores signal; when $k$ is larger, the additional neighbors lie where $K_h(s - x^\star)$ is negligible and contribute only additional noise. The optimal $k$ therefore depends on how the generator distributes its excess mass across synthetic points.
    \item \emph{Bandwidth-scale matching.} The signal term $\Psi(h, \rho, x^\star)$ is a kernel-overlap integral between $K_h$ and the $\phi_\rho$, and is non-negligible only in a neighborhood of $x^\star$ on the scale of $\rho$. The bandwidth $h$ controls the resolution at which the KDE recovers the local density ratio $p_{R \cup \{x^\star\}}/p_R$ in this neighborhood: $h \gg \rho$ oversmooths and washes out the perturbation from adding $x^\star$, while $h \ll \rho$ produces an estimate of $\hat{p}_R$ that adds noise to the score. Because the attack signal is local, estimators with explicit control over a local smoothing scale are better attack instruments than estimators optimized for global density-estimation accuracy.

\end{enumerate}

We study the empirical behavior of these quantities under controlled conditions in Section~\ref{sec:simulation}, and connect them to Gen-LRA's performance on real generators in Section~\ref{sec:experiments}.

\subsection{Implementation Choices}
\label{sec:implementation}
 
The theoretical results above motivate each component of the Gen-LRA implementation (full pseudocode is given in Alg.~\ref{alg:gen_lra}).
 
\paragraph{Surrogate model}
We use Gaussian KDEs as the surrogate density estimator. This choice is motivated by Theorem~\ref{thm:closedform}: the closed-form expansion depends on Gaussian-kernel algebra, making Gen-LRA with KDEs analytically tractable and computationally cheap to run. KDEs are widely available and, per the remark following Theorem~\ref{thm:meangap}, their locally structured bias profile aligns with the task of detecting memorization at scale $\rho \approx h$. We include an ablation and discussion where we study replacing KDEs with a flow-based deep learning method (see Section~\ref{subsec:deeplearningdensityestimation}), but find KDE empirically outperforms it. We set the bandwidth using Silverman's rule per results we find in Section~\ref{sec:simulation}. An ablation on encoding strategies for heterogeneous tabular data appears in Appendix~\ref{app:encoding}; we that ordinal encoding for categorical variables outperforms other methods and thus choose it as a default.
 
\paragraph{Choice of k}
The choice of $k$ governs a bias-variance trade-off (Section~\ref{sec:what_measures}), and it plays different roles in the two settings Gen-LRA is used in. A realistic adversary cannot tune $k$ against ground-truth membership labels and must commit to a single value; the ablation in Appendix~\ref{app:k} shows that $k = 10$ is robust across datasets and generators, and we adopt it as the default. A privacy auditor, by contrast, has access to the true membership labels by construction and seeks the tightest possible bound on the generator's vulnerability. For a given $x^\star$, the dominant cost is the $2k$ KDE evaluations against reference sets of size $m$ and $m+1$, giving a per-query cost of $\mathcal{O}(k \cdot m \cdot d)$. Computing scores for all $k' \in \{1, \ldots, k\}$ requires no additional KDE evaluations: the per-neighbor log-density-ratio terms can be summed cumulatively in $\mathcal{O}(k)$ time. Sweeping $k$ exhaustively up to $k_{\max}$ is therefore asymptotically equivalent to a single evaluation at $k_{\max}$, at cost $\mathcal{O}(k_{\max} \cdot m \cdot d)$. An auditor can thus report the strongest configuration over $k$, while our adversary defaults to $k = 10$.

\paragraph{Decision threshold}
The membership prediction rule~\eqref{eq:membership_prediction} requires a decision threshold $\gamma$. In principle $\gamma$ is tunable for a specific application, but $\hat{\mathcal{I}}(x^\star; R, S) > 0$ implies some degree of local overfitting to $x^\star$ and serves as a natural threshold.

\begin{algorithm}[H]
\caption{Gen-LRA}
\label{alg:gen_lra}
\begin{algorithmic}[1]
\Require Test set $X_{\text{test}}$, synthetic set $S$, reference set $R$, locality $k$
\Ensure Membership scores $\{a_{x^\star}\}_{x^\star \in X_{\text{test}}}$
\State $\hat{p}_R \gets \textsc{FitKDE}(R)$ \Comment{Surrogate on reference only}
\For{$x^\star \in X_{\text{test}}$}
    \State $\hat{p}_{R \cup \{x^\star\}} \gets \textsc{FitKDE}(R \cup \{x^\star\})$ \Comment{Augmented surrogate}
    \State $S_k(x^\star) \gets \textsc{kNN}(S, x^\star, k)$ \Comment{Localize to $k$-NN synthetic points}
    \State $a_{x^\star} \gets \displaystyle\sum_{s \in S_k(x^\star)} \left[\log \hat{p}_{R \cup \{x^\star\}}(s) - \log \hat{p}_R(s)\right]$
\EndFor
\State \Return $\{a_{x^\star}\}_{x^\star \in X_{\text{test}}}$
\end{algorithmic}
\end{algorithm}
\section{Simulation Study}
\label{sec:simulation}
 
Before benchmarking Gen-LRA on real tabular generators (Section~\ref{sec:experiments}), we conduct a controlled simulation study to directly examine the structural properties of Theorems~\ref{thm:closedform} and~\ref{thm:meangap}. The motivation is that real generators confound three questions: whether the attack works, under what conditions it works, and whether the parameters $(\epsilon, \rho)$ are meaningful in practice. A simulation lets us address all three under controlled conditions, and additionally study how the linearity, bandwidth-scale and localization properties from Section~\ref{sec:when_works} manifest empirically. 

\subsection{Simulation Design}
\label{sec:simulation_setup}
 
We directly instantiate the local overfitting model of Definition~\ref{def:localoverfit}. Concretely, we fix a population distribution $P = \mathcal{N}(0, I_d)$ and sample $T$, a held-out non-member set, and $R$ independently from $P$, each of size $n = 500$. To generate the synthetic set $S$, each of $n$ synthetic points is drawn as
\begin{equation}
s_i \sim \begin{cases}
P & \text{with probability } 1 - \epsilon, \\
\mathcal{N}(x_j, \rho^2 I_d), \; x_j \sim \text{Unif}(T) & \text{with probability } \epsilon.
\end{cases}
\end{equation}
This directly implements Equation~\ref{eq:overfitmodel} with $\phi_\rho(\cdot; x_j) = \mathcal{N}(\cdot; x_j, \rho^2 I_d)$, making $\epsilon$ and $\rho$ controllable configurations rather than latent properties of a trained model. We use $d = 40$, consistent with moderate-dimensional tabular data.
 
\paragraph{Attacks.} We evaluate Gen-LRA against two common threat-model-compatible baselines: DOMIAS~\citep{vanbreugel2023membership}, which computes a pointwise density ratio $\frac{\hat{p}_S(x^\star)}{\hat{p}_R(x^\star)}$, and DCR~\citep{ganleaks}, which scores test points by distance to the nearest synthetic neighbor. We additionally include two variants of Gen-LRA, one with $k = 1$ (single-neighbor localization) and one with $k = 10$ (the default used in our benchmark experiments). Including both variants lets us isolate the role of the localization parameter $k$ in shaping the attack's global and tail behavior.
 
\paragraph{Experimental grid.} We compute $\epsilon \in \{0, 0.25, 0.5, 0.75, 1.0\}$ and $\rho \in \{0.1, 0.2, \ldots, 1.7\}$ over 10 random seeds per grid point. For each $(\epsilon, \rho, \text{seed})$ combination, we compute attack performance using Area Under the Curve (AUC) and True Positive Rate at the Fixed False Positive Rate of 0 (TPR@FPR=0) \citep{Carlini2021MembershipIA} by scoring all members and held-out non-members. This yields a measurement of both the global attack performance and the calibration of the classifier under various privacy leakage settings.
 
\subsection{Simulation Results}
\label{sec:simulation_results}
\begin{figure}[t]
  \centering
  \includegraphics[width=\linewidth]{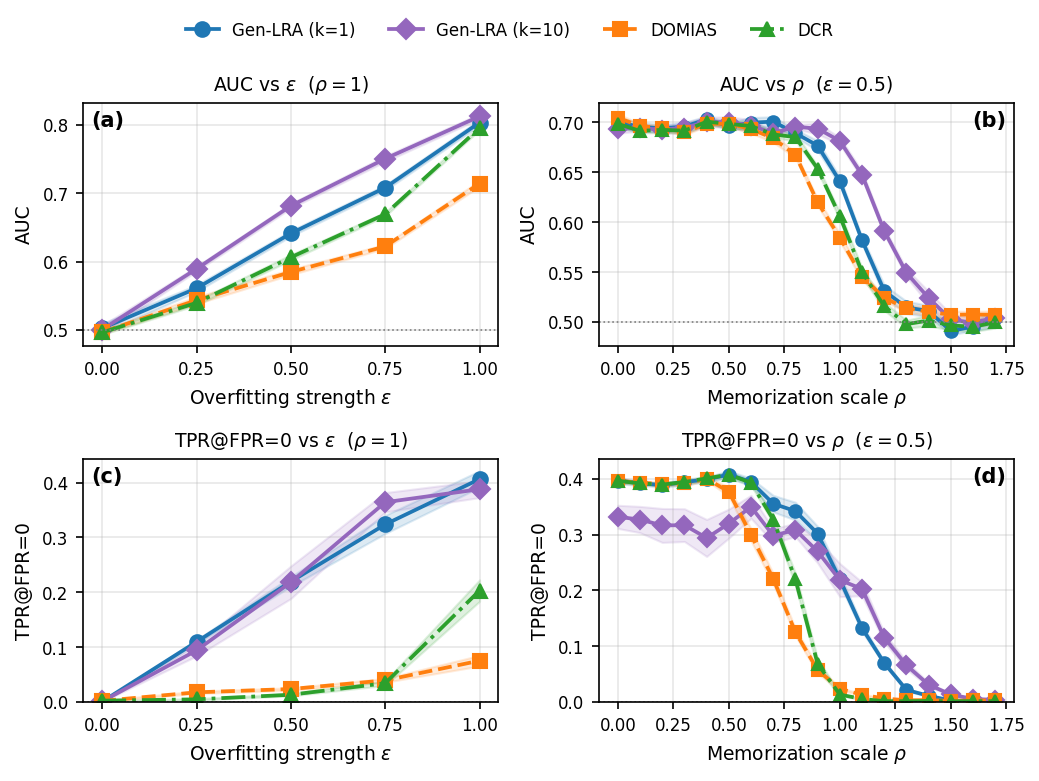}
\caption{Simulation results across attacks and metrics. \textbf{(a, c)} AUC and TPR@FPR=0 versus $\epsilon$ at $\rho = 1$. All attacks scale linearly with $\epsilon$; Gen-LRA achieves the largest slope on both metrics. \textbf{(b, d)} AUC and TPR@FPR=0 versus $\rho$ at $\epsilon = 0.5$. At small $\rho$ all attacks perform comparably; at moderate $\rho$ Gen-LRA dominates while baselines collapse; at large $\rho$ all attacks approach random. The locality parameter $k$ controls how Gen-LRA aggregates evidence across synthetic neighbors: $k=1$ recovers the full TPR@FPR=0 signal under near-exact memorization, where evidence concentrates on a single neighbor, while $k=10$ outperforms it at moderate $\rho$, where the signal is dispersed across multiple neighbors.}  
\Description{A two-by-two grid of line plots comparing four attacks across two metrics and two experimental sweeps. The four attacks shown in each panel are Gen-LRA with k equal to 1, Gen-LRA with k equal to 10, DOMIAS, and DCR. The top row reports area under the ROC curve and the bottom row reports true positive rate at false positive rate zero. The left column varies the overfitting strength epsilon from 0 to 1 at fixed memorization scale rho equal to 1; in panels a and c, all four attacks rise approximately linearly with epsilon, with Gen-LRA at k equal to 10 reaching the highest values, around 0.81 AUC and 0.40 true positive rate at the rightmost point. The right column varies the memorization scale rho from 0 to 1.75 at fixed overfitting strength epsilon equal to 0.5; in panels b and d, all attacks start near comparable values at small rho, with DOMIAS, DCR, and Gen-LRA at k equal to 1 collapsing toward random performance as rho increases past about 0.7, while Gen-LRA at k equal to 10 sustains higher performance through moderate rho before all four attacks converge to random at large rho. Shaded bands around each curve indicate variability across random seeds.}
\label{fig:simulation}
\end{figure}

 Figure~\ref{fig:simulation} presents four primary results. The top row reports mean AUC; the bottom row reports mean TPR@FPR=0 with respective error bars. The left column varies $\epsilon$ at fixed memorization scale; the right column varies $\rho$ at fixed overfitting strength.
\subsubsection{Linear Dependence on $\epsilon$}
Panels (a) and (c) report AUC and TPR@FPR=0 as functions of $\epsilon$ at fixed memorization scale $\rho = 1$. Both metrics scale linearly with $\epsilon$ for all four attacks, validating Theorem~\ref{thm:meangap}'s prediction that the mean score gap is linear in the overfitting strength. Gen-LRA achieves the largest slope on both metrics, with $k = 10$ reaching AUC $\approx 0.81$ at $\epsilon = 1$ on panel (a) and $k = 1$ and $k = 10$ both reaching TPR@FPR=0 $\approx 0.40$ on panel (c). 

The advantage over baselines is substantially larger for TPR@FPR =0 than AUC: at $\epsilon = 1$, Gen-LRA's TPR@FPR=0 is roughly $2\times$ that of DCR and $4\times$ that of DOMIAS, whereas the corresponding AUC ratios are smaller. This reflects the structural advantage of Gen-LRA's likelihood-ratio formulation. The theorem does not predict particular slope magnitudes, but the empirical ordering matches the structural argument in Section~\ref{sec:what_measures}: aggregating evidence over $k$ synthetic neighbors yields a stronger linear response than the single-point attacks DOMIAS and DCR.

\subsubsection{Localization and Three Regimes of Memorization}
Panels (b) and (d) sweep $\rho$ at fixed $\epsilon = 0.5$. Recall that $\rho$ is the scale of noise added to a training point when producing a memorized synthetic point: small $\rho$ means synthetic points are nearly exact copies of training points, large $\rho$ means they are heavily perturbed. The curves divide into three regimes that correspond to different types of empirical memorization.
 
\textit{Near-exact memorization} ($\rho \lesssim 0.7$). Synthetic points memorized from training data are nearly identical to their source records. For a member $x^\star \in T$, this produces a single synthetic point $s_1$ that is a near-copy of $x^\star$, and adding $x^\star$ to the reference set creates a sharp, narrow peak in $\hat{p}_{R \cup \{x^\star\}}$ centered at $x^\star$. The likelihood ratio $\hat{p}_{R \cup \{x^\star\}}(s)/\hat{p}_R(s)$ carries useful signal only for synthetic points near $x^*$; elsewhere the two density estimates are nearly identical and the ratio is uninformative. DOMIAS, DCR, and Gen-LRA $k=1$ target this peak. All three capture the full membership signal and achieve comparable AUC ($\approx 0.69$) and TPR@FPR=0 ($\approx 0.40$). Gen-LRA $k=10$ underperforms on TPR@FPR=0 here as included additional synthetic neighbors are unrelated to $x^\star$ and their contributions effectively add noise to the score.

\textit{Approximate memorization} ($\rho \in [0.7, 1.5]$). Synthetic points retain a recognizable relationship to training points but are spread out enough that no single synthetic point is a reliable copy of any given training point. The peak in $\hat{p}_{R \cup \{x^\star\}}$ is still localized to $x^\star$, but probability mass in $S$ is no longer concentrated at $x^*$: instead, several synthetic points sit near $x^\star$, each carrying partial evidence. DOMIAS and DCR collapse in this regime. DOMIAS's single-point density ratio loses signal because the synthetic data is no longer concentrated at $x^\star$, and DCR loses signal because the closest synthetic point to a member is no longer reliably closer than the closest synthetic point to a non-member. Gen-LRA $k=1$ also degrades but at a lower rate than DCR and DOMIAS. Gen-LRA $k=10$ however, aggregates evidence across all the relevant synthetic neighbors and dominates on both AUC and TPR@FPR=0.

\textit{Diffuse memorization} ($\rho \gtrsim 1.5$). Synthetic points are perturbed heavily. No local feature of $S$ distinguishes regions near training points from regions near non-training points, and all four attacks collapse to random performance.
\begin{figure}[t]
    \centering
    \includegraphics[width=\linewidth]{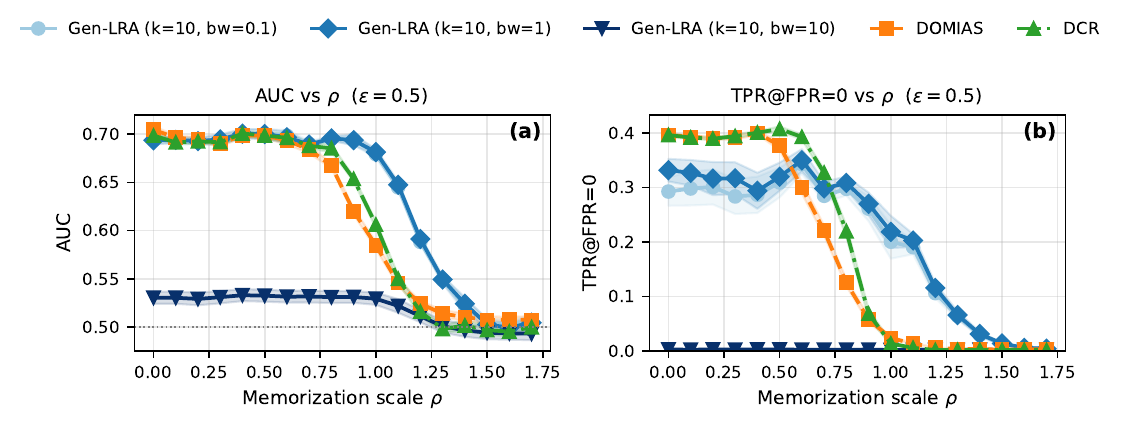}
    \caption{AUC and TPR@FPR=0 versus memorization scale $\rho$ for Gen-LRA $k=10$ at three bandwidths ($0.1\times$, $1\times$, and $10\times$ Silverman's rule), with DOMIAS and DCR shown for reference. Narrow and default bandwidths produce nearly indistinguishable curves; the wide bandwidth ($10\times$) oversmooths the perturbation from $x^\star$ and collapses attack performance to near-random.}
    \label{fig:bandwidth}
    \Description{Two side-by-side line plots showing AUC and true positive rate at false positive rate zero versus memorization scale rho. Five curves per panel: Gen-LRA with k=10 at three bandwidths (0.1 times, 1 times, and 10 times Silverman's rule), plus DOMIAS and DCR. The 0.1 times and 1 times Gen-LRA curves overlap nearly identically and dominate the other curves at small to moderate rho before declining toward random at large rho. The 10 times Gen-LRA curve sits flat near random performance across all rho values. DOMIAS and DCR decline toward random faster than the working Gen-LRA configurations.}
\end{figure}
\subsubsection{The Role of Bandwidth}
Figure~\ref{fig:bandwidth} repeats the $\rho$ sweep from panels (b) and (d) for Gen-LRA at three bandwidths: $0.1\times$, $1\times$, and $10\times$ Silverman's rule. The narrow ($0.1\times$) and default ($1\times$) bandwidths produce nearly indistinguishable curves on both AUC and TPR@FPR=0, indicating that Gen-LRA is robust to under-smoothing in this regime. The wide bandwidth ($10\times$) collapses Gen-LRA's performance to near-random AUC and near-zero TPR@FPR=0 across all $\rho$. This corresponds with item~(3) of Section~\ref{sec:when_works}: when $h \gg \rho$, the KDE oversmooths the perturbation introduced by adding $x^\star$ to $R$, washing out the local density ratio that Gen-LRA relies on; when $h \ll \rho$, the estimate is noisier but empirically retains the structural signal that the score depends on.

\begin{table*}[t]
\caption{Attack ranking performance across evaluation metrics over all 1{,}525 runs. Higher Top-k values are better, while lower mean rank indicates stronger performance. Gen-LRA achieves the highest Top-1 rate on every metric at more than $2\times$ that of any baseline.}
\label{tab:rank}
\centering
\small
\begin{tabular}{lllllll}
\toprule
& & \multirow{2}{*}{} & \multicolumn{4}{c}{TPR at Fixed FPR} \\
\cmidrule(lr){4-7}
Method & Statistic & AUC & 0.0 & 0.001 & 0.01 & 0.1 \\
\midrule

\multirow[t]{3}{*}{Gen LRA $k=10$}
& Top-1 & \textbf{38.6\%} & \textbf{42.1\%} & \textbf{41.5\%} & \textbf{39.8\%} & \textbf{46.2\%} \\
& Top-3 & \textbf{71.7\%} & \textbf{78.5\%} & \textbf{75.6\%} & \textbf{73.1\%} & \textbf{76.7\%} \\
& Mean Rank (std) & \textbf{2.83 (2.12)} & \textbf{2.79 (2.05)} & \textbf{2.76 (2.02)} & \textbf{2.78 (2.04)} & \textbf{2.54 (2.0)} \\
\cline{1-7}

\multirow[t]{3}{*}{Classifier}
& Top-1 & 16.0\% & 9.6\% & 9.1\% & 11.6\% & 9.6\% \\
& Top-3 & 45.7\% & 35.4\% & 31.9\% & 35.2\% & 35.7\% \\
& Mean Rank (std) & 4.29 (2.57) & 5.22 (2.16) & 5.04 (2.36) & 4.79 (2.47) & 4.83 (2.48) \\
\cline{1-7}

\multirow[t]{3}{*}{DCR}
& Top-1 & 9.0\% & 17.4\% & 15.0\% & 12.2\% & 8.3\% \\
& Top-3 & 31.2\% & 44.1\% & 37.4\% & 37.9\% & 32.2\% \\
& Mean Rank (std) & 4.93 (2.34) & 4.67 (2.3) & 4.77 (2.44) & 4.63 (2.39) & 4.86 (2.3) \\
\cline{1-7}

\multirow[t]{3}{*}{DCR-Diff}
& Top-1 & 12.0\% & 18.4\% & 16.4\% & 16.3\% & 16.3\% \\
& Top-3 & 43.5\% & 60.3\% & 51.8\% & 50.0\% & 46.2\% \\
& Mean Rank (std) & 4.11 (2.21) & 3.78 (2.04) & 4.05 (2.35) & 4.2 (2.49) & 4.2 (2.48) \\
\cline{1-7}

\multirow[t]{3}{*}{DOMIAS}
& Top-1 & 10.0\% & 12.7\% & 11.8\% & 11.3\% & 11.8\% \\
& Top-3 & 36.8\% & 52.1\% & 43.4\% & 36.8\% & 41.0\% \\
& Mean Rank (std) & 4.42 (2.19) & 4.22 (2.08) & 4.56 (2.45) & 5.03 (2.67) & 4.63 (2.59) \\
\cline{1-7}

\multirow[t]{3}{*}{DPI}
& Top-1 & 10.7\% & 2.4\% & 1.9\% & 4.0\% & 6.5\% \\
& Top-3 & 38.9\% & 13.0\% & 12.4\% & 20.6\% & 29.9\% \\
& Mean Rank (std) & 4.5 (2.34) & 6.79 (1.51) & 5.94 (1.95) & 5.18 (1.93) & 4.97 (2.22) \\

\cline{1-7}

\multirow[t]{3}{*}{LOGAN}
& Top-1 & 12.7\% & 16.2\% & 13.8\% & 11.3\% & 11.0\% \\
& Top-3 & 28.7\% & 48.9\% & 41.7\% & 30.8\% & 32.1\% \\
& Mean Rank (std) & 5.51 (2.79) & 4.35 (2.2) & 4.71 (2.53) & 5.3 (2.59) & 5.19 (2.62) \\
\cline{1-7}

\multirow[t]{3}{*}{Local Neighborhood}
& Top-1 & 2.4\% & 6.6\% & 5.8\% & 6.7\% & 4.4\% \\
& Top-3 & 16.0\% & 32.9\% & 30.3\% & 29.9\% & 20.8\% \\
& Mean Rank (std) & 5.79 (1.99) & 5.4 (2.09) & 5.26 (2.28) & 5.11 (2.3) & 5.41 (2.14) \\
\cline{1-7}

\multirow[t]{3}{*}{MC}
& Top-1 & 4.2\% & 12.5\% & 11.3\% & 8.5\% & 6.3\% \\
& Top-3 & 22.5\% & 43.8\% & 35.7\% & 30.3\% & 25.3\% \\
& Mean Rank (std) & 5.61 (2.3) & 4.71 (2.14) & 4.91 (2.41) & 5.11 (2.41) & 5.32 (2.35) \\
\bottomrule
\end{tabular}
\end{table*}

\section{Experiments}
\label{sec:experiments}

The simulation study in Section~\ref{sec:simulation} shows that different attacks and hyperparameter settings perform differently across memorization regimes. This raises a methodological concern: a small benchmark risks producing results that are artifacts of which regimes happen to dominate the selection, biasing the comparison toward attacks suited to those particular conditions. We therefore construct a large and diverse benchmark to assess Gen-LRA's effectiveness across a broad distribution of synthetic datasets: 35 tabular datasets sampled from OpenML \citep{oml-benchmarking-suites}, 9 tabular generative architectures, 9 membership inference attacks, and 5 random seeds. This yields 1{,}525 unique synthetic datasets and over 10{,}000 unique attack runs, spanning a wide range of memorization conditions that approximate what practitioners could encounter in deployment.
\subsection{Benchmark Setup}
\label{sec:benchmarking}

For each dataset, following standard tabular synthetic data evaluation practice~\citep{synthcity}, we randomly partition the data without replacement into a training set $T$, holdout set $H$, and reference set $R$ in an 80/10/10 ratio. The generator is trained on $T$ and produces a synthetic dataset $S$ of size $|T|$. MIAs are then evaluated on a balanced test set composed of $\min(1000, |H|)$ members sampled from $T$ and an equal number of non-members sampled from $H$. Full details on datasets, MIAs, and generators are provided in Appendix~\ref{app:benchmark_details}.

Gen-LRA and DOMIAS rely on density estimation, which we implement using Gaussian Kernel Density Estimation (KDE) with Silverman's rule for bandwidth selection. We find KDE outperforms deep learning based density estimators on this task for Gen-LRA (Section~\ref{subsec:deeplearningdensityestimation}). Because KDE handles one hot encoded categorical data poorly, we use ordinal encoding for these attacks; an ablation across Principle Component Analysis and Variational Autoencoder based encodings appears in Appendix~\ref{app:encoding} and shows that ordinal encoding performs best. For all other attacks, numeric features are scaled and categorical features one hot encoded. 

\subsection{Baselines}
\label{sec:baselines}

We compare Gen-LRA with a fixed $k=10$ against eight competing MIAs that operate under compatible threat models: LOGAN, MC, DCR, DCR-Diff, Classifier, Local Neighborhood, DOMIAS, and DPI~\citep{Hayes2017LOGANMI, Hilprecht2019MonteCA, ganleaks, houssiau2022tapas, vanbreugel2023membership, ward2024dataplagiarismindexcharacterizing}. 

We evaluate across nine tabular generators spanning the major architectural families: PATEGAN ($\epsilon=1$) for differentially private generation; Ads-GAN, CTGAN, and TVAE for adversarial training; Normalizing Flows for likelihood-based generation; ARF for tree-based generation; Tab-DDPM and TabSyn for diffusion-based generation; and RealTabFormer (RTF) for transformer-based generation~\citep{Ankan2015, yoon2018pategan, yoon2020anonymization, Xu2019ModelingTD, durkan2019neural, pmlr-v206-watson23a, tabddpm, tabsyn,solatorio2023realtabformer}. For RealTabFormer and TabSyn we use the original implementations with default hyperparameters; for all other architectures we use the default Synthcity~\citep{synthcity} implementations and settings.

All experiments were conducted on an AWS G5.2xlarge EC2 instance. The main benchmark required approximately 500 hours of compute, with an additional 80 hours used for the deep learning density estimation experiments described in Section~\ref{subsec:deeplearningdensityestimation}.
\section{Experiment Results}
\label{sec:benchmark_results}
 
We summarize benchmark performance along two complementary axes: aggregate ranking across all 1{,}525 synthetic datasets, and performance restricted to the top 100 runs by each metric. The aggregate ranking measures how often each attack is the strongest available choice across diverse synthetic datasets a practitioner may encounter. The top-100 view measures attack performance in the synthetic datasets with the greatest amount of detected leakage.
 
\subsection{Aggregate Ranking}
\label{sec:results_aggregate}
 
An adversary attacking an unknown generator must commit to a single attack as they have no ground truth labels, so a relevant question is which attack most often delivers the strongest signal across a diverse sample of synthetic datasets. Table~\ref{tab:rank} reports, for each attack and metric, the fraction of runs in which the attack ranks first (Top-1), the fraction of runs in which it ranks in the top three (Top-3), and the mean rank across all 1{,}525 runs. The Top-1 rate answers this question directly, and Gen-LRA is the dominant choice. Gen-LRA achieves the highest Top-1 rate on every metric (38.6\% on AUC, 42.1\% on TPR@FPR=0, and 46.2\% on TPR@FPR=0.1), more than $2\times$ the rate of any baseline. 

The closest competitors are DCR-Diff at 18.4\% on TPR@FPR=0 and Classifier at 16.0\% on AUC; no baseline exceeds a 19\% Top-1 rate on any metric. Gen-LRA's margin widens at higher false-positive rates, reaching $2.8\times$ the next-best attack at TPR@FPR=0.1, indicating that the benefit of aggregating evidence over $k$ synthetic neighbors is most pronounced at operating points permitting a moderate false-positive budget. The supporting Top-3 rates (71.7\%--78.5\% for Gen-LRA, against $\leq 60.3\%$ for any baseline) and mean ranks (2.54--2.83 for Gen-LRA, against $\geq 3.78$ for any baseline) confirm that this dominance is consistent.
\begin{table}[t]
\caption{Mean performance with standard deviations across the top 100 runs for each metric. Bold indicates the highest value in each column. Despite outliers that inflate attacks that target near-memorization, Gen-LRA achieves the highest mean on every metric.}

\centering
\small
\setlength{\tabcolsep}{4pt}
\begin{tabular}{lcccc}
\toprule
& \multirow{2}{*}{} & \multicolumn{3}{c}{TPR at Fixed FPR} \\
\cmidrule(lr){3-5}
Method & AUC & 0.0 & 0.01 & 0.1 \\
\midrule
Gen-LRA & \textbf{0.589 (0.03)} & \textbf{0.051 (0.03)} & \textbf{0.053 (0.03)} & \textbf{0.211 (0.05)} \\
Classifier & 0.557 (0.05) & 0.012 (0.02) & 0.021 (0.02) & 0.144 (0.06) \\
DCR & 0.579 (0.04) & 0.029 (0.04) & 0.040 (0.04) & 0.175 (0.06) \\
DCR-Diff & 0.570 (0.04) & 0.037 (0.04) & 0.044 (0.04) & 0.171 (0.05) \\
DOMIAS & 0.546 (0.04) & 0.024 (0.02) & 0.024 (0.02) & 0.140 (0.04) \\
DPI & 0.526 (0.03) & 0.001 (0.00) & 0.013 (0.01) & 0.114 (0.02) \\
LOGAN & 0.493 (0.02) & 0.013 (0.01) & 0.015 (0.01) & 0.105 (0.03) \\
Local N. & 0.538 (0.04) & 0.013 (0.02) & 0.024 (0.02) & 0.132 (0.04) \\
MC & 0.548 (0.04) & 0.016 (0.02) & 0.024 (0.02) & 0.134 (0.05) \\
\bottomrule
\end{tabular}
\label{tab:mean}
\end{table}

\subsection{Performance on the Most Vulnerable Runs}
\label{sec:results_top100}
Aggregate ranking measures an adversary's preference for an attack; we now turn to attack performance for runs that exhibited the highest detected leakage. Table~\ref{tab:mean} reports each metric's mean and standard deviation over the top 100 runs by that metric, the regime in which the adversary's attack choice has the largest practical consequence. Gen-LRA achieves the highest mean AUC (0.59) and the highest mean TPR at every false-positive rate, with the largest gap at low FPR: Gen-LRA's mean TPR@FPR=0 of 0.05 is approximately $1.7\times$ that of DCR-Diff and $2.5\times$ that of DOMIAS.
 
We additionally report medians and interquartile ranges in the Appendix Table~\ref{tab:median_iqr}, which shows that the closest baseline performance is outlier driven. DCR and DCR-Diff have median TPR@FPR =0 values (0.011 and 0.024) well below their means, with IQRs touching zero. Gen-LRA's median TPR@FPR=0 of 0.055 exceeds its own mean, its IQR is bounded away from zero ($[0.034, 0.067]$), and its median AUC of 0.594 exceeds the third quartile of every baseline. 

These outliers are concentrated on datasets generated by RealTabFormer, which accounts for 42 of the top 100 runs by AUC. Recently, \cite{Ward2026StringMem} documented that RealTabFormer often produces near-exact copies of training records. The simulation in Section~\ref{sec:simulation_results} identifies this small $\rho$ regime as the one in which distance based baselines are most competitive with Gen-LRA at the fixed $k=10$ used in our benchmark, implying that the top 100 runs overrepresent the regime least favorable to Gen-LRA, yet Gen-LRA leads on every metric in both mean and median.
 \begin{figure}
    \centering
    \includegraphics[width=\linewidth]{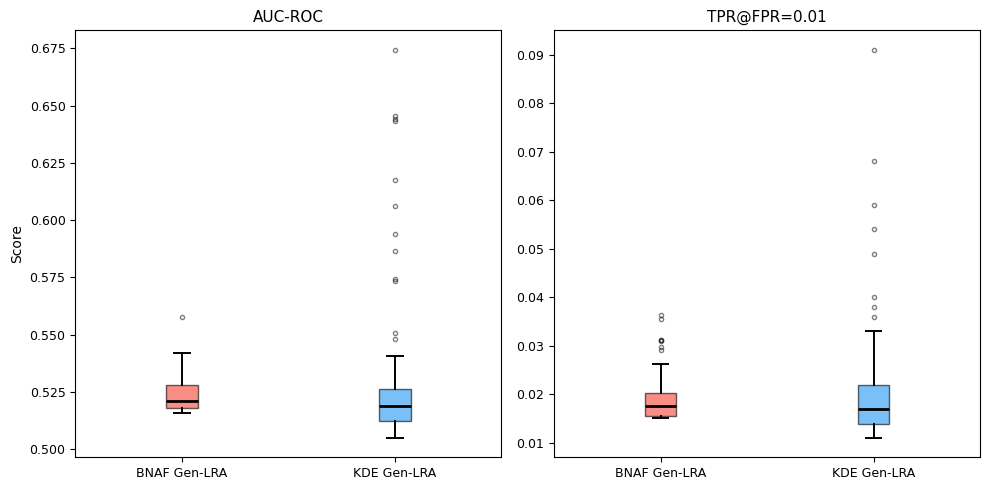}
 \caption{Distribution of Gen-LRA performance across benchmark runs using BNAF and KDE as the surrogate density estimator, on AUC-ROC (left) and TPR@FPR=0.01 (right). BNAF produces a tighter distribution of scores but fails to detect the strongest leakage signals: its upper tail is bounded well below KDE's, whose long upper tail captures the runs on which the most vulnerable generators leak training-set membership.}
    \Description{Two side-by-side boxplots comparing Gen-LRA performance using BNAF versus KDE as the density-estimation surrogate. The left panel reports AUC-ROC and the right panel reports true positive rate at false positive rate 0.01. In each panel, the BNAF box is narrow and centered just above 0.5 AUC and 0.02 TPR, with whiskers and outliers all bounded below 0.56 AUC and 0.04 TPR. The KDE box has a similar median but a much longer upper tail of outliers extending to roughly 0.67 AUC and 0.09 TPR, corresponding to runs in which KDE detects strong privacy leakage that BNAF fails to identify. The pattern is consistent across both metrics: BNAF is more stable but flat near random, while KDE captures the high-leakage runs that matter for privacy auditing.}
\label{fig:bnafvskde}
\end{figure}

\begin{figure*}
    \centering
    \includegraphics[width=.75\linewidth]{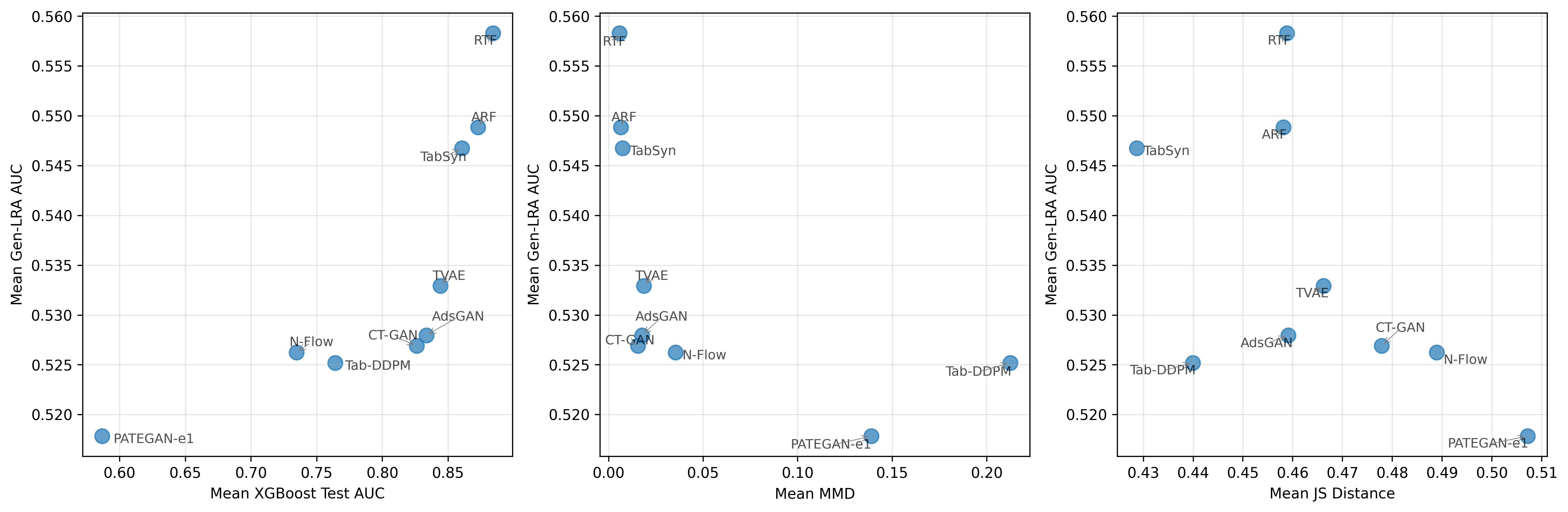}
\caption{Mean Gen-LRA AUC plotted against three quality measures across generators: downstream utility (XGBoost test AUC under train-on-synthetic-test-on-real), maximum mean discrepancy, and Jensen-Shannon distance. }
\Description{Three side-by-side scatter plots, each showing mean Gen-LRA AUC on the y-axis (ranging roughly 0.52 to 0.56) plotted against a different synthetic data quality measure on the x-axis. From left to right, the x-axes are XGBoost test AUC under train-on-synthetic-test-on-real (downstream utility), mean maximum mean discrepancy, and mean Jensen-Shannon distance. Each point is labeled with a generator name: PATE-GAN at epsilon equal to 1, Tab-DDPM, CTGAN, Ads-GAN, TVAE, N-Flow, ARF, TabSyn, and RealTabFormer. PATE-GAN sits in the worst-quality, lowest-leakage corner of all three panels, with the lowest XGBoost test AUC, the highest MMD, and the highest JS distance, paired with the lowest Gen-LRA AUC near 0.518. RealTabFormer, ARF, and TabSyn cluster in the opposite corner of each panel, with the highest utility and lowest MMD and JS distances, paired with the highest Gen-LRA AUC values around 0.555. The remaining generators fall between these two extremes, producing a clear trend in which higher synthetic data quality corresponds to higher detected privacy leakage.}
\label{fig:fidelity_utility}
\end{figure*}

\subsection{Deep Learning Density Estimation}
\label{subsec:deeplearningdensityestimation}
Gen-LRA estimates the likelihood of high dimensional data, a setting in which Gaussian Kernel Density Estimation (KDE) is typically outperformed by modern density estimators on metrics such as average negative log likelihood and negative evidence lower bound~\citep{bnaf19, pmlr-v162-wen22c}. It is therefore worth examining whether replacing KDE with a more expressive estimator improves attack performance. Following~\cite{vanbreugel2023membership}, we perform an ablation (details in Appendix~\ref{app:dl}) with Block Neural Autoregressive Flows (BNAF) \citep{bnaf19} as the surrogate density estimator for Gen-LRA.

Figure~\ref{fig:bnafvskde} shows that KDE matches or outperforms BNAF on both AUC and TPR@FPR=0.01, with the largest gap on the generators that exhibit the most extreme privacy leakage. While BNAF produces a tighter distribution of scores across runs, this stability comes from a consistent failure to detect the strongest leakage signals: BNAF's upper tail tops out near 0.56 AUC, whereas KDE reaches above 0.65 on the same generators. The bandwidth analysis in Section~\ref{sec:when_works} explains this result: the Gen-LRA signal is governed by the kernel overlap term $\Psi(h, \rho, x^\star)$, which requires the surrogate's local smoothing scale to match the memorization scale $\rho$. KDE controls this scale explicitly through its bandwidth. BNAF however optimizes for global density accuracy and exposes no analogous parameter, so its implicit smoothing scale is set by global training dynamics rather than matched to $\rho$. The result is a surrogate that is more expressive globally but less faithful in the local neighborhood where Gen-LRA's signal is found.

A second issue compounds the first: Gen-LRA compares two surrogates fit on $R$ and $R \cup \{x^\star\}$, and recovering the influence of a single point requires the difference between them to reflect that point rather than training stochasticity. KDE is deterministic given its bandwidth and data, so the two fits differ only in the contribution of $x^\star$. BNAF and other modern density estimators are trained by stochastic optimization, and the two fits differ in initialization, batch order, and optimizer state, which adds noise to the influence estimate of Gen-LRA. We therefore default to KDE for Gen-LRA, and recommend it to practitioners on the basis of empirical performance and substantially lower computational cost.

\subsection{Privacy-Utility Trade-off}
\label{sec:results_utility}
 
The preceding results compare attacks against each other; we now ask whether the privacy leakage detected by Gen-LRA tracks the quality of the synthetic data being audited. Figure~\ref{fig:fidelity_utility} plots mean Gen-LRA AUC against three quality measures averaged across models for each benchmark run: XGBoost test AUC under a train-on-synthetic-test-on-real protocol (downstream utility), maximum mean discrepancy and Jensen-Shannon distance between real and synthetic samples (distributional fidelity). Lower MMD and JS values indicate higher fidelity, while higher XGBoost test AUC indicates higher utility.
 
PATE-GAN at $\epsilon=1$ occupies the worst-quality corner of all three panels, with the lowest utility (XGBoost AUC $\approx 0.59$) and the worst fidelity on both MMD and JS, and also the lowest Gen-LRA AUC ($\approx 0.518$). This is the expected behavior of differentially private synthesis: the privacy guarantee that limits Gen-LRA's signal also limits the synthetic data's resemblance to the population. At the opposite end, RealTabFormer, ARF, and TabSyn achieve the highest utility and the strongest fidelity scores while exhibiting the highest Gen-LRA AUC values in the benchmark. These generators produce synthetic data that supports downstream learning effectively, but the same fidelity that makes the synthetic data useful renders it more vulnerable to No-Box MIAs.
\section{Discussion}
\subsection{Interpreting the Benchmark}

Theorem~\ref{thm:closedform} characterizes the Gen-LRA score as a sum of localized density ratios over the $k$ synthetic neighbors of $x^\star$, and Theorem~\ref{thm:meangap} shows that the expected score gap between members and non-members is governed by the kernel-overlap term $\Psi(h, \rho, x^\star)$ between the surrogate's bandwidth and the generator's memorization scale. The simulation in Section~\ref{sec:simulation} instantiates this score in three regimes the theory makes precise. Under near-exact memorization, single-point statistics already capture the membership signal and aggregation over $k$ neighbors only adds noise. Under approximate memorization, signal is dispersed across multiple synthetic points and aggregation recovers what single-point statistics miss. Under diffuse memorization, no local statistic separates members from non-members, a regime in which all No-Box attacks must fail. Extending the model to capture non-uniform overfitting across training points, for instance, by allowing $\epsilon$ and $\rho$ to vary with the local density of the population, is a natural refinement.

The distribution of overfitting behavior across modern tabular generators is not known a priori, and prior MIA evaluations have rarely been broad enough to characterize it. Most published tabular generative model MIA works span a handful of datasets and a small set of generator architectures, leaving open the question of whether reported performance generalizes beyond the conditions of the benchmark itself. Our evaluation spans 1{,}525 synthetic datasets across 9 generator families and 35 datasets, covering the architectural and statistical diversity a practitioner could realistically encounter.

Across this distribution, Gen-LRA achieves the highest Top-1 rate on every metric at more than $2\times$ that of any baseline, indicating that the approximate-memorization regime predicted by Theorem~\ref{thm:meangap} to favor Gen-LRA is well-represented in practice. RealTabFormer is the most prominent exception, with known near-exact memorization behavior that dominates the top-100 runs by AUC and accounts for the outlier-driven means observed for distance-based baselines. Even within this subset, Gen-LRA achieves the highest mean and median on every metric.

These numbers also understate Gen-LRA's true power. We fix $k = 10$ across all 1{,}525 runs rather than tuning it per dataset or per generator, but a privacy auditor has access to ground-truth membership labels by construction, and the cost analysis in Section~\ref{sec:implementation} shows that sweeping $k$ up to $k_{\max}$ is asymptotically equivalent to a single evaluation at $k_{\max}$. An auditor using Gen-LRA can therefore report the strongest configuration over $k$ at no additional cost, yielding strictly tighter bounds than the fixed-$k$ numbers reported here. The benchmark results should be read as a conservative estimate of what Gen-LRA delivers in practice. 

The flip side is also informative: at fixed $k = 10$, Gen-LRA is \emph{not} the top-ranked attack on roughly 60\% of runs. No single attack, Gen-LRA included, is uniformly optimal across datasets and generator architectures in our benchmark. Whether a uniformly optimal No-Box MIA exists is an important open question with direct consequences for how synthetic data should be audited in practice.

\subsection{No-Box Auditing Estimates Realistic Privacy Leakage}

A common objection to No-Box MIAs is that they are weaker than shadow-box or query-based alternatives, and therefore underestimate privacy risk. This framing conflates two distinct questions. Privacy leakage is not a property of a model alone — it is a property of a model conditioned on a threat model. The relevant question for a practitioner releasing synthetic data is not "what is the maximum leakage achievable by any conceivable adversary?" but "what leakage is achievable by an adversary I should plausibly defend against?"

Shadow and white-box attacks assume the adversary knows the generator's architecture or has direct access. A defender following standard release practice, withholding model details and releasing only the synthetic dataset, eliminates the information these attacks depend on. The leakage they measure is therefore not the leakage the release exposes, but the leakage that would be exposed under a counterfactual disclosure regime the defender did not adopt. The No-Box setting, by contrast, corresponds to the release itself: the adversary sees only the synthetic dataset and auxiliary population data they could plausibly obtain. Less conservative threat models remain essential for other auditing problems — verifying differential privacy guarantees, for instance, requires the strongest attacks available regardless of deployment realism \citep{annamalai2024whatwanttheoryalone}. But sample-level auditing under realistic adversaries is equally important, and progress in the No-Box regime, both theoretical and empirical, is what gives practitioners the tools to assess the privacy of the artifacts they actually release.
 
\subsection{Limitations}
Gen-LRA's theoretical guarantees and empirical performance come with three limitations, each suggesting a direction for future work.

\paragraph{Scope of the local overfitting model.} Definition~\ref{def:localoverfit} characterizes the generator's output density as the population density plus additive excess mass concentrated within distance $\rho$ of each training point, and Theorem~\ref{thm:meangap} provides guarantees only for generators whose output density admits this decomposition with $\epsilon > 0$ and $\rho$ comparable to the surrogate's bandwidth. The model further idealizes $\phi_\rho$ and $\epsilon$ as uniform across training points, whereas in practice memorization is non-uniform: rare records, outliers, and points in low-density regions are typically memorized more strongly than the rest. The model is nonetheless rich enough to capture the three regimes that distinguish Gen-LRA's behavior from single-point baselines (Section~\ref{sec:simulation}), and its predictions are consistent with Gen-LRA's empirical performance on real generators.

\paragraph{Per-target hyperparameter sub-optimality.} The non-uniformity of memorization carries through to Gen-LRA itself: a single choice of bandwidth $h$ and locality $k$ is unlikely to be simultaneously optimal for every $x^\star$, and per-target attack power varies with the local memorization geometry around each test point. Our work does not disentangle this per-target variation. That said, the cost analysis in Section~\ref{sec:implementation} establishes that adaptive per-target configurations are computationally feasible, making this a natural and tractable direction for future work rather than an obstacle.

\paragraph{Surrogate scaling.} KDE is known to scale poorly with dimensionality, and the empirical observation that KDE outperforms BNAF in our benchmark relies on the moderate dimensions ($d \lesssim 100$ after encoding) of the OpenML datasets we evaluate. On substantially higher-dimensional tabular data, neither KDE nor BNAF may be adequate, and Gen-LRA's surrogate component would require revisiting. The density-ratio-estimation literature~\cite{sugiyama2012density, kanamori2009least} establishes that estimating a ratio directly is more sample-efficient than estimating two densities and dividing them, suggesting an attractive alternative to the two surrogate fits in Algorithm~\ref{alg:gen_lra}. Adapting these methods to Gen-LRA is non-trivial: standard direct-ratio estimators assume two independent samples drawn from distinct distributions, whereas the two distributions whose ratio we require are induced by reference sets that differ by a single point. Closing this gap likely requires new estimators tailored to the influence-function structure of the problem, which we view as a promising direction for scaling Gen-LRA to higher-dimensional regimes. 

\section{Conclusion}

Our results suggest that the dominant No-Box membership inference attacks for synthetic data release underestimate the privacy risk of the artifacts they evaluate. DOMIAS tests \emph{the wrong hypothesis}, evaluating where synthetic density exceeds reference density rather than whether a specific point influenced the synthetic distribution; and distance-based methods such as DCR detect leakage only when memorization is near-exact. Both classes underestimate in the approximate-memorization regime, which our benchmark shows is where most modern tabular generators likely operate. A practitioner concluding that a release is private on the basis of these attacks is relying on measurements whose inductive biases are matched to a specific overfitting phenomena that the generator may not exhibit.

Gen-LRA addresses this gap with an attack that is both principled and practical. The likelihood-ratio influence-function formulation directly tests the membership hypothesis, the closed-form characterization (Theorem~\ref{thm:closedform}) makes explicit what the attack measures, and the mean-score-gap result (Theorem~\ref{thm:meangap}) provides testable predictions about when it succeeds, predictions we validate in a controlled simulation and that align with empirical performance across 1{,}525 synthetic datasets spanning nine generators. Gen-LRA achieves the highest Top-1 rate on every metric at more than $2\times$ that of any baseline, and corresponding highest mean values in the top 100 highest privacy leakage runs.

The framework we develop suggests several directions for future work. Adaptive bandwidth and locality selection on a per-target basis would likely increase attack performance, particularly for the long-tail records that are memorized most strongly. Replacing the two separate density estimates in Algorithm~\ref{alg:gen_lra} with a direct density-ratio estimator tailored to the single-point-difference setting offers a route to scaling Gen-LRA beyond the moderate-dimensional regimes where KDE remains effective, and would likely improve performance on our benchmark as well. Extending the local-overfitting model and the influence-function framing to settings where synthetic tabular data release is becoming common such as longitudinal records is a natural next step. Finally, Gen-LRA's broad but non-uniform dominance leaves open whether a uniformly optimal No-Box MIA exists---a question we view as central to the privacy auditing literature, since such an attack would let practitioners directly certify a synthetic dataset's privacy under the threat model rather than relying on guarantees from defensive frameworks.

\section{Ethical Considerations}
This work proposes a membership inference attack against synthetic tabular data, which is itself a re-identification method. We address the resulting ethical considerations along three dimensions: the risks the attack poses, the benefits it provides, and the steps we have taken to ensure that the latter outweigh the former.

\textit{Risks.} Adversaries who can infer whether an individual's record was used to train a generative model pose direct privacy risks to that individual, particularly in domains such as healthcare, finance, and the social sciences where sensitive personal data is routinely used. A synthetic dataset that fails to obfuscate membership information can enable re-identification of training-set participants, undermining the privacy guarantees that motivate synthetic data release in the first place. Gen-LRA, by improving the state of the art in No-Box membership inference, raises the ceiling on what such an adversary can achieve.

\textit{Benefits.} The same capability that creates risk also enables stronger privacy auditing. Practitioners releasing synthetic datasets currently rely on similarity-based heuristics that have been shown to be inadequate proxies for privacy, or on attacks whose threat-model assumptions do not match realistic release scenarios. Gen-LRA gives auditors a principled, computationally tractable tool to assess sample-level privacy leakage under the threat model that actually corresponds to public release. Without attacks of this kind, defenders cannot measure the privacy of the artifacts they release.

\textit{Steps taken to minimize harm.} We evaluate Gen-LRA exclusively on publicly available datasets from OpenML, which contain no personally identifying information beyond what their original releasers have already chosen to publish. We do not target any deployed synthetic data release, nor do we attempt to re-identify individuals in any real-world dataset. The generative models we attack are trained by us on these public datasets specifically for the purpose of evaluating Gen-LRA, so no third party's privacy claims are challenged by our experiments. We release our code openly to support reproducibility and to enable defenders to audit their own releases, which we view as the primary use case. Responsible disclosure does not apply here, as no specific deployed system is implicated; the vulnerability we characterize is structural to overfitting in tabular generative models and has been documented in prior work.

We believe adversarial work of this kind is essential to the development of trustworthy synthetic data release. Deferring the publication of stronger attacks would leave practitioners auditing with tools that systematically underestimate privacy risk, which we view as a net riskier outcome.

\bibliographystyle{ACM-Reference-Format}
\bibliography{main}

@String{Computing = "Computing" }

@String{Computer = "{IEEE} Computer" }

@String{Springer = "Springer-Verlag" }

@misc{golob2024privacyvulnerabilitiesmarginalsbasedsynthetic,
      title={Privacy Vulnerabilities in Marginals-based Synthetic Data}, 
      author={Steven Golob and Sikha Pentyala and Anuar Maratkhan and Martine De Cock},
      year={2024},
      eprint={2410.05506},
      archivePrefix={arXiv},
      primaryClass={cs.CR},
      url={https://arxiv.org/abs/2410.05506}, 
}

@inproceedings{ganleaks, series={CCS ’20},
   title={GAN-Leaks: A Taxonomy of Membership Inference Attacks against Generative Models},
   url={http://dx.doi.org/10.1145/3372297.3417238},
   DOI={10.1145/3372297.3417238},
   booktitle={Proceedings of the 2020 ACM SIGSAC Conference on Computer and Communications Security},
   publisher={ACM},
   author={Chen, Dingfan and Yu, Ning and Zhang, Yang and Fritz, Mario},
   year={2020},
   month=oct, collection={CCS ’20} }

@INPROCEEDINGS {Shokri,
author = {R. Shokri and M. Stronati and C. Song and V. Shmatikov},
booktitle = {2017 IEEE Symposium on Security and Privacy (SP)},
title = {Membership Inference Attacks Against Machine Learning Models},
year = {2017},
volume = {},
issn = {2375-1207},
pages = {3-18},
doi = {10.1109/SP.2017.41},
url = {https://doi.ieeecomputersociety.org/10.1109/SP.2017.41},
publisher = {IEEE Computer Society},
address = {Los Alamitos, CA, USA},
month = {may}
}

@inproceedings{Sablayrolles2019WhiteboxVB,
  title={White-box vs black-box: Bayes optimal strategies for membership inference},
  author={Sablayrolles, Alexandre and Douze, Matthijs and Schmid, Cordelia and Ollivier, Yann and J{\'e}gou, Herv{\'e}},
  booktitle={Proceedings of the 36th International Conference on Machine Learning},
  volume={97},
  pages={5558--5567},
  year={2019},
  month=jun,
  publisher={PMLR},
  address={Long Beach, CA, USA},
  series={Proceedings of Machine Learning Research}
}

@article{Hayes2017LOGANMI,
  title={LOGAN: Membership Inference Attacks Against Generative Models},
  author={Jamie Hayes and Luca Melis and George Danezis and Emiliano De Cristofaro},
  journal={Proceedings on Privacy Enhancing Technologies},
  year={2017},
  volume={2019},
  pages={133 - 152},
  url={https://api.semanticscholar.org/CorpusID:52211986}
}

@article{Hilprecht2019MonteCA,
  title={Monte Carlo and Reconstruction Membership Inference Attacks against Generative Models},
  author={Benjamin Hilprecht and Martin H{\"a}rterich and Daniel Bernau},
  journal={Proceedings on Privacy Enhancing Technologies},
  year={2019},
  volume={2019},
  pages={232 - 249},
  url={https://api.semanticscholar.org/CorpusID:199546273}
}

@InProceedings{vanbreugel2023membership,
  title = 	 {Membership Inference Attacks against Synthetic Data through Overfitting Detection},
  author =       {van Breugel, Boris and Sun, Hao and Qian, Zhaozhi and van der Schaar, Mihaela},
  booktitle = 	 {Proceedings of The 26th International Conference on Artificial Intelligence and Statistics},
  pages = 	 {3493--3514},
  year = 	 {2023},
  editor = 	 {Ruiz, Francisco and Dy, Jennifer and van de Meent, Jan-Willem},
  volume = 	 {206},
  series = 	 {Proceedings of Machine Learning Research},
  month = 	 {25--27 Apr},
  publisher =    {PMLR},
  url = 	 {https://proceedings.mlr.press/v206/breugel23a.html}
}

@inbook{Meeus_2024,
   title={Achilles’ Heels: Vulnerable Record Identification in Synthetic Data Publishing},
   ISBN={9783031514760},
   ISSN={1611-3349},
   url={http://dx.doi.org/10.1007/978-3-031-51476-0_19},
   DOI={10.1007/978-3-031-51476-0_19},
   booktitle={Lecture Notes in Computer Science},
   publisher={Springer Nature Switzerland},
   author={Meeus, Matthieu and Guepin, Florent and Creţu, Ana-Maria and de Montjoye, Yves-Alexandre},
   year={2024},
   pages={380–399} }

@misc{ganev2023inadequacy,
      title={On the Inadequacy of Similarity-based Privacy Metrics: Reconstruction Attacks against "Truly Anonymous Synthetic Data''}, 
      author={Georgi Ganev and Emiliano De Cristofaro},
      year={2023},
      eprint={2312.05114},
      archivePrefix={arXiv},
      primaryClass={cs.CR}
}

@article{bnaf19,
  title={Block Neural Autoregressive Flow},
  author={De Cao, Nicola and
          Titov, Ivan and
          Aziz, Wilker},
  journal={35th Conference on Uncertainty in Artificial Intelligence (UAI19)},
  year={2019}
}

@misc{tabddpm,
      title={TabDDPM: Modelling Tabular Data with Diffusion Models}, 
      author={Akim Kotelnikov and Dmitry Baranchuk and Ivan Rubachev and Artem Babenko},
      year={2022},
      eprint={2209.15421},
      archivePrefix={arXiv},
      primaryClass={cs.LG}
}

@inproceedings{
tabsyn,
title={Mixed-Type Tabular Data Synthesis with Score-based Diffusion in Latent Space},
author={Hengrui Zhang and Jiani Zhang and Zhengyuan Shen and Balasubramaniam Srinivasan and Xiao Qin and Christos Faloutsos and Huzefa Rangwala and George Karypis},
booktitle={The Twelfth International Conference on Learning Representations},
year={2024},
url={https://openreview.net/forum?id=4Ay23yeuz0}
}

@misc{adult,
  author       = {Becker,Barry and Kohavi,Ronny},
  title        = {{Adult}},
  year         = {1996},
  howpublished = {UCI Machine Learning Repository},
  note         = {{DOI}: https://doi.org/10.24432/C5XW20}
}

@INPROCEEDINGS {9458927,
author = {S. Takagi and T. Takahashi and Y. Cao and M. Yoshikawa},
booktitle = {2021 IEEE 37th International Conference on Data Engineering (ICDE)},
title = {P3GM: Private High-Dimensional Data Release via Privacy Preserving Phased Generative Model},
year = {2021},
volume = {},
issn = {},
pages = {169-180},
doi = {10.1109/ICDE51399.2021.00022},
url = {https://doi.ieeecomputersociety.org/10.1109/ICDE51399.2021.00022},
publisher = {IEEE Computer Society},
address = {Los Alamitos, CA, USA},
month = {apr}
}

@InProceedings{pmlr-v162-wen22c,
  title = 	 {Random Forest Density Estimation},
  author =       {Wen, Hongwei and Hang, Hanyuan},
  booktitle = 	 {Proceedings of the 39th International Conference on Machine Learning},
  pages = 	 {23701--23722},
  year = 	 {2022},
  editor = 	 {Chaudhuri, Kamalika and Jegelka, Stefanie and Song, Le and Szepesvari, Csaba and Niu, Gang and Sabato, Sivan},
  volume = 	 {162},
  series = 	 {Proceedings of Machine Learning Research},
  month = 	 {17--23 Jul},
  publisher =    {PMLR},
  url = 	 {https://proceedings.mlr.press/v162/wen22c.html}
}

@InProceedings{10.1007/978-3-642-28914-9_19,
author="Gupta, Anupam
and Roth, Aaron
and Ullman, Jonathan",
editor="Cramer, Ronald",
title="Iterative Constructions and Private Data Release",
booktitle="Theory of Cryptography",
year="2012",
publisher="Springer Berlin Heidelberg",
address="Berlin, Heidelberg",
pages="339--356",
isbn="978-3-642-28914-9"
}

@inproceedings{10.5555/2095116.2095131,
author = {Hardt, Moritz and Rothblum, Guy N. and Servedio, Rocco A.},
title = {Private data release via learning thresholds},
year = {2012},
publisher = {Society for Industrial and Applied Mathematics},
address = {USA},

booktitle = {Proceedings of the Twenty-Third Annual ACM-SIAM Symposium on Discrete Algorithms},
pages = {168–187},
numpages = {20},
location = {Kyoto, Japan},
series = {SODA '12}
}

@article{WANG2022306,
title = {Differentially private SGD with non-smooth losses},
journal = {Applied and Computational Harmonic Analysis},
volume = {56},
pages = {306-336},
year = {2022},
issn = {1063-5203},
doi = {https://doi.org/10.1016/j.acha.2021.09.001},
url = {https://www.sciencedirect.com/science/article/pii/S1063520321000841},
author = {Puyu Wang and Yunwen Lei and Yiming Ying and Hai Zhang},

}

@inproceedings{Abadi_2016, series={CCS’16},
   title={Deep Learning with Differential Privacy},
   url={http://dx.doi.org/10.1145/2976749.2978318},
   DOI={10.1145/2976749.2978318},
   booktitle={Proceedings of the 2016 ACM SIGSAC Conference on Computer and Communications Security},
   publisher={ACM},
   author={Abadi, Martin and Chu, Andy and Goodfellow, Ian and McMahan, H. Brendan and Mironov, Ilya and Talwar, Kunal and Zhang, Li},
   year={2016},
   month=oct, collection={CCS’16} }

@article{yoon2020anonymization,  title={Anonymization through data synthesis using generative adversarial networks (ads-gan)},  author={Yoon, Jinsung and Drumright, Lydia N and Van Der Schaar, Mihaela},  journal={IEEE journal of biomedical and health informatics},  volume={24},  number={8},  pages={2378--2388},  year={2020},  publisher={IEEE}}

@article{solatorio2023realtabformer,  title={Realtabformer: Generating realistic relational and tabular data using transformers},  author={Solatorio, Aivin V and Dupriez, Olivier},  journal={arXiv preprint arXiv:2302.02041},  year={2023}}

@article{park2018data,
author = {Park, Noseong and Mohammadi, Mahmoud and Gorde, Kshitij and Jajodia, Sushil and Park, Hongkyu and Kim, Youngmin},
title = {Data synthesis based on generative adversarial networks},
year = {2018},
issue_date = {June 2018},
publisher = {VLDB Endowment},
volume = {11},
number = {10},
issn = {2150-8097},
url = {https://doi.org/10.14778/3231751.3231757},
doi = {10.14778/3231751.3231757},
journal = {Proc. VLDB Endow.},
month = jun,
pages = {1071–1083},
numpages = {13}
}

@article{platzer2021holdout,  title={Holdout-based empirical assessment of mixed-type synthetic data},  author={Platzer, Michael and Reutterer, Thomas},  journal={Frontiers in big Data},  volume={4},  pages={679939},  year={2021},  publisher={Frontiers Media SA}}

@inproceedings{Xu2019ModelingTD,
  title={Modeling Tabular data using Conditional GAN},
  author={Lei Xu and Maria Skoularidou and Alfredo Cuesta-Infante and Kalyan Veeramachaneni},
  booktitle={Neural Information Processing Systems},
  year={2019},
  url={https://api.semanticscholar.org/CorpusID:195767064}
}

@inbook{durkan2019neural,
author = {Durkan, Conor and Bekasov, Artur and Murray, Iain and Papamakarios, George},
title = {Neural spline flows},
year = {2019},
publisher = {Curran Associates Inc.},
address = {Red Hook, NY, USA},
booktitle = {Proceedings of the 33rd International Conference on Neural Information Processing Systems},
articleno = {675},
numpages = {12}
}

@inproceedings{
yoon2018pategan,
title={{PATE}-{GAN}: Generating Synthetic Data with Differential Privacy Guarantees},
author={Jinsung Yoon and James Jordon and Mihaela van der Schaar},
booktitle={International Conference on Learning Representations},
year={2019},
url={https://openreview.net/forum?id=S1zk9iRqF7},
}

@article {PMID:32167919,
	Title = {Anonymization Through Data Synthesis Using Generative Adversarial Networks (ADS-GAN)},
	Author = {Yoon, Jinsung and Drumright, Lydia N and van der Schaar, Mihaela},
	DOI = {10.1109/jbhi.2020.2980262},
	Number = {8},
	Volume = {24},
	Month = {August},
	Year = {2020},
	Journal = {IEEE journal of biomedical and health informatics},
	ISSN = {2168-2194},
	Pages = {2378—2388},
	URL = {https://doi.org/10.1109/jbhi.2020.2980262},
}

@inproceedings{Ankan2015,
  series = {SciPy},
  title = {pgmpy: Probabilistic Graphical Models using Python},
  ISSN = {2575-9752},
  url = {http://dx.doi.org/10.25080/Majora-7b98e3ed-001},
  DOI = {10.25080/majora-7b98e3ed-001},
  booktitle = {Proceedings of the Python in Science Conference},
  publisher = {SciPy},
  author = {Ankan,  Ankur and Panda,  Abinash},
  year = {2015},
  collection = {SciPy}
}

@InProceedings{pmlr-v206-watson23a,
  title = 	 {Adversarial Random Forests for Density Estimation and Generative Modeling},
  author =       {Watson, David S. and Blesch, Kristin and Kapar, Jan and Wright, Marvin N.},
  booktitle = 	 {Proceedings of The 26th International Conference on Artificial Intelligence and Statistics},
  pages = 	 {5357--5375},
  year = 	 {2023},
  editor = 	 {Ruiz, Francisco and Dy, Jennifer and van de Meent, Jan-Willem},
  volume = 	 {206},
  series = 	 {Proceedings of Machine Learning Research},
  month = 	 {25--27 Apr},
  publisher =    {PMLR},
  url = 	 {https://proceedings.mlr.press/v206/watson23a.html}
}

@inproceedings{lu2019empirical,
author = {Lu, Pei-Hsuan and Wang, Pang-Chieh and Yu, Chia-Mu},
title = {Empirical Evaluation on Synthetic Data Generation with Generative Adversarial Network},
year = {2019},
isbn = {9781450361903},
publisher = {Association for Computing Machinery},
address = {New York, NY, USA},
url = {https://doi.org/10.1145/3326467.3326474},
doi = {10.1145/3326467.3326474},
booktitle = {Proceedings of the 9th International Conference on Web Intelligence, Mining and Semantics},
articleno = {16},
numpages = {6},
location = {Seoul, Republic of Korea},
series = {WIMS2019}
}

@inproceedings{yale2019assessing,
author = {Yale, Andrew and Dash, Saloni and Dutta, Ritik and Guyon, Isabelle and Pavao, Adrien and Bennett, Kristin P.},
title = {Assessing privacy and quality of synthetic health data},
year = {2019},
isbn = {9781450371841},
publisher = {Association for Computing Machinery},
address = {New York, NY, USA},
url = {https://doi.org/10.1145/3359115.3359124},
doi = {10.1145/3359115.3359124},
booktitle = {Proceedings of the Conference on Artificial Intelligence for Data Discovery and Reuse},
articleno = {8},
numpages = {4},
location = {Pittsburgh, Pennsylvania},
series = {AIDR '19}
}

@InProceedings{zhao2021ctab,
  title = 	 {CTAB-GAN: Effective Table Data Synthesizing},
  author =       {Zhao, Zilong and Kunar, Aditya and Birke, Robert and Chen, Lydia Y.},
  booktitle = 	 {Proceedings of The 13th Asian Conference on Machine Learning},
  pages = 	 {97--112},
  year = 	 {2021},
  editor = 	 {Balasubramanian, Vineeth N. and Tsang, Ivor},
  volume = 	 {157},
  series = 	 {Proceedings of Machine Learning Research},
  month = 	 {17--19 Nov},
  publisher =    {PMLR},
  url = 	 {https://proceedings.mlr.press/v157/zhao21a.html}
}

@article{guillaudeux2023patient,
author = {Guillaudeux, Morgan and Rousseau, Olivia and Petot, Julien and Bennis, Zineb and Dein, Charles-Axel and Goronflot, Thomas and Karakachoff, Matilde and Limou, Sophie and Vince, Nicolas and Wargny, Matthieu and Gourraud, Pierre-Antoine},
year = {2022},
month = {05},
pages = {},
title = {Patient-centric synthetic data generation, no reason to risk re-identification in the analysis of biomedical pseudonymised data},
doi = {10.21203/rs.3.rs-1674043/v1}
}

@article{liu2023tabular,
author = {Liu, Tongyu and Fan, Ju and Li, Guoliang and Tang, Nan and Du, Xiaoyong},
title = {Tabular data synthesis with generative adversarial networks: design space and optimizations},
year = {2023},
issue_date = {Mar 2024},
publisher = {Springer-Verlag},
address = {Berlin, Heidelberg},
volume = {33},
number = {2},
issn = {1066-8888},
url = {https://doi.org/10.1007/s00778-023-00807-y},
doi = {10.1007/s00778-023-00807-y},
journal = {The VLDB Journal},
month = {aug},
pages = {255–280},
numpages = {26}
}

@misc{ims1,
   title={Truly Anonymous Synthetic Data -- Evolving Legal Definitions and
  Technologies (Part II)},
   url={https://mostly.ai/blog/truly-anonymous-synthetic-data-legal-definitions-part-ii/},
   author={Mostly AI},
   year={2020} }

@misc{ims2,
   title={How to implement data privacy? A conversation with Klaudius
  Kalcher},
   url={https://mostly.ai/data-democratization-podcast/how-to-implement-data-privacy/},
   author={Mostly AI},
   year={2021} }

@misc{synthcity,
  doi = {10.48550/ARXIV.2301.07573},
  url = {https://arxiv.org/abs/2301.07573},
  author = {Qian, Zhaozhi and Cebere, Bogdan-Constantin and van der Schaar, Mihaela},
  title = {Synthcity: facilitating innovative use cases of synthetic data in different data modalities},
  year = {2023},
  copyright = {Creative Commons Attribution 4.0 International}
}

@article{houssiau2022tapas,  title={Tapas: a toolbox for adversarial privacy auditing of synthetic data},  author={Houssiau, Florimond and Jordon, James and Cohen, Samuel N and Daniel, Owen and Elliott, Andrew and Geddes, James and Mole, Callum and Rangel-Smith, Camila and Szpruch, Lukasz},  journal={arXiv preprint arXiv:2211.06550},  year={2022}}

@article{Carlini2021MembershipIA,
  title={Membership Inference Attacks From First Principles},
  author={Nicholas Carlini and Steve Chien and Milad Nasr and Shuang Song and A. Terzis and Florian Tram{\`e}r},
  journal={2022 IEEE Symposium on Security and Privacy (SP)},
  year={2021},
  pages={1897-1914},
  url={https://api.semanticscholar.org/CorpusID:244920593}
}

@article{ward2024dataplagiarismindexcharacterizing,
      title={Data Plagiarism Index: Characterizing the Privacy Risk of Data-Copying in Tabular Generative Models}, 
    journal = "KDD- Generative AI Evaluation Workshop",
      author={Joshua Ward and Chi-Hua Wang and Guang Cheng},
      year={2024},
      eprint={2406.13012},
      archivePrefix={arXiv},
      primaryClass={cs.LG},
      url={https://arxiv.org/abs/2406.13012}, 
}

@inproceedings{ye2022,
author = {Ye, Jiayuan and Maddi, Aadyaa and Murakonda, Sasi Kumar and Bindschaedler, Vincent and Shokri, Reza},
title = {Enhanced Membership Inference Attacks against Machine Learning Models},
year = {2022},
isbn = {9781450394505},
publisher = {Association for Computing Machinery},
address = {New York, NY, USA},
url = {https://doi.org/10.1145/3548606.3560675},
doi = {10.1145/3548606.3560675},
booktitle = {Proceedings of the 2022 ACM SIGSAC Conference on Computer and Communications Security},
pages = {3093–3106},
numpages = {14},
location = {Los Angeles, CA, USA},
series = {CCS '22}
}

@misc{zarifzadeh2024,
      title={Low-Cost High-Power Membership Inference Attacks}, 
      author={Sajjad Zarifzadeh and Philippe Liu and Reza Shokri},
      year={2024},
      eprint={2312.03262},
      archivePrefix={arXiv},
      primaryClass={stat.ML},
      url={https://arxiv.org/abs/2312.03262}, 
}

@INPROCEEDINGS {Yeom,
author = {S. Yeom and I. Giacomelli and M. Fredrikson and S. Jha},
booktitle = {2018 IEEE 31st Computer Security Foundations Symposium (CSF)},
title = {Privacy Risk in Machine Learning: Analyzing the Connection to Overfitting},
year = {2018},
volume = {},
issn = {2374-8303},
pages = {268-282},
doi = {10.1109/CSF.2018.00027},
url = {https://doi.ieeecomputersociety.org/10.1109/CSF.2018.00027},
publisher = {IEEE Computer Society},
address = {Los Alamitos, CA, USA},
month = {jul}
}

@INPROCEEDINGS{Long,
  author={Long, Yunhui and Wang, Lei and Bu, Diyue and Bindschaedler, Vincent and Wang, Xiaofeng and Tang, Haixu and Gunter, Carl A. and Chen, Kai},
  booktitle={2020 IEEE European Symposium on Security and Privacy (EuroS\&P)}, 
  title={A Pragmatic Approach to Membership Inferences on Machine Learning Models}, 
  year={2020},
  volume={},
  number={},
  pages={521-534},
  doi={10.1109/EuroSP48549.2020.00040}}

@inproceedings{Song2020SystematicEO,
  title={Systematic Evaluation of Privacy Risks of Machine Learning Models},
  author={Liwei Song and Prateek Mittal},
  booktitle={USENIX Security Symposium},
  year={2020},
  url={https://api.semanticscholar.org/CorpusID:214623088}
}

@inproceedings{
watson2022on,
title={On the Importance of Difficulty Calibration in Membership Inference Attacks},
author={Lauren Watson and Chuan Guo and Graham Cormode and Alexandre Sablayrolles},
booktitle={International Conference on Learning Representations},
year={2022},
url={https://openreview.net/forum?id=3eIrli0TwQ}
}

@inproceedings {groundhog,
author = {Theresa Stadler and Bristena Oprisanu and Carmela Troncoso},
title = {Synthetic Data {\textendash} Anonymisation Groundhog Day},
booktitle = {31st USENIX Security Symposium (USENIX Security 22)},
year = {2022},
isbn = {978-1-939133-31-1},
address = {Boston, MA},
pages = {1451--1468},
url = {https://www.usenix.org/conference/usenixsecurity22/presentation/stadler},
publisher = {USENIX Association},
month = aug
}

@article{hampel1974influence,
 ISSN = {01621459, 1537274X},
 URL = {http://www.jstor.org/stable/2285666},
 author = {Frank R. Hampel},
 journal = {Journal of the American Statistical Association},
 number = {346},
 pages = {383--393},
 publisher = {[American Statistical Association, Taylor \& Francis, Ltd.]},
 title = {The Influence Curve and Its Role in Robust Estimation},
 urldate = {2025-05-15},
 volume = {69},
 year = {1974}
}

@book{cook1986residuals,
  title={Residuals and Influence in Regression},
  author={Cook, R.D. and Weisberg, S.},
  series={Monographs on statistics and applied probability},
  year={1986},
  publisher={Chapman and Hall}
}

@InProceedings{pmlr-v70-koh17a,
  title = 	 {Understanding Black-box Predictions via Influence Functions},
  author =       {Pang Wei Koh and Percy Liang},
  booktitle = 	 {Proceedings of the 34th International Conference on Machine Learning},
  pages = 	 {1885--1894},
  year = 	 {2017},
  editor = 	 {Precup, Doina and Teh, Yee Whye},
  volume = 	 {70},
  series = 	 {Proceedings of Machine Learning Research},
  month = 	 {06--11 Aug},
  publisher =    {PMLR},
  url = 	 {https://proceedings.mlr.press/v70/koh17a.html}
}

@book{sugiyama2012density,
  title     = {Density Ratio Estimation in Machine Learning},
  author    = {Sugiyama, Masashi and Suzuki, Taiji and Kanamori, Takafumi},
  year      = {2012},
  publisher = {Cambridge University Press},
  address   = {Cambridge},
  doi       = {10.1017/CBO9781139035613}
}

@article{kanamori2009least,
  title   = {A Least-squares Approach to Direct Importance Estimation},
  author  = {Kanamori, Takafumi and Hido, Shohei and Sugiyama, Masashi},
  journal = {Journal of Machine Learning Research},
  volume  = {10},
  pages   = {1391--1445},
  year    = {2009}
}

@misc{annamalai2024whatwanttheoryalone,
      title={"What do you want from theory alone?" Experimenting with Tight Auditing of Differentially Private Synthetic Data Generation}, 
      author={Meenatchi Sundaram Muthu Selva Annamalai and Georgi Ganev and Emiliano De Cristofaro},
      year={2024},
      eprint={2405.10994},
      archivePrefix={arXiv},
      primaryClass={cs.CR},
      url={https://arxiv.org/abs/2405.10994}, 
}

@article{oml-benchmarking-suites,

      title={OpenML Benchmarking Suites}, 

      author={Bernd Bischl and Giuseppe Casalicchio and Matthias Feurer and Frank Hutter and Michel Lang and Rafael G. Mantovani and Jan N. van Rijn and Joaquin Vanschoren},

      year={2019},

      journal={arXiv:1708.03731v2 [stat.ML]}

}

@misc{Ward2026StringMem,
      title={When Tables Leak: Attacking String Memorization in LLM-Based Tabular Data Generation}, 
      author={Joshua Ward and Bochao Gu and Chi-Hua Wang and Guang Cheng},
      year={2025},
      eprint={2512.08875},
      archivePrefix={arXiv},
      primaryClass={cs.LG},
      url={https://arxiv.org/abs/2512.08875}, 
}

%
\appendix

\section{Proofs and Supporting Results}
\label{app:proofs}

This appendix provides proofs for the three theoretical results of Section~\ref{sec:genlra}.

\subsection{Proof of Theorem~\ref{thm:invariance} (Invariance)}
\label{app:invariance_proof}

\begin{proof}
Let $g: \mathcal{X} \to \mathcal{X}$ be a continuously differentiable invertible function with Jacobian $J(x) = \frac{dg}{dx}(x)$, where $|J(x)| \neq 0$ for all $x \in \mathcal{X}$. For sets $A \subseteq \mathcal{X}$, write $\tilde{A} = g(A)$, and for a density $p$ let $\tilde{p}$ denote the density induced on the transformed space.

By the change-of-variables formula for probability densities,
\begin{equation}
\tilde{p}(g(A)) \;=\; \frac{p(A)}{|J(A)|},
\end{equation}
where $|J(A)|$ denotes the product of Jacobian determinants evaluated on $A$ (treating $J$ as acting pointwise on a set of samples).

Applying this to the conditional density in the numerator of the log-likelihood ratio:
\begin{align}
\tilde{p}(\tilde{S} \mid \tilde{R} \cup \{\tilde{x}^\star\})
\;&=\; \frac{\tilde{p}(\tilde{S}, \tilde{R} \cup \{\tilde{x}^\star\})}{\tilde{p}(\tilde{R} \cup \{\tilde{x}^\star\})} \\
\;&=\; \frac{p(S, R \cup \{x^\star\}) / |J(S, R \cup \{x^\star\})|}{p(R \cup \{x^\star\}) / |J(R \cup \{x^\star\})|}.
\end{align}
Similarly for the denominator:
\begin{align}
\tilde{p}(\tilde{S} \mid \tilde{R})
\;&=\; \frac{\tilde{p}(\tilde{S}, \tilde{R})}{\tilde{p}(\tilde{R})}
\;=\; \frac{p(S, R) / |J(S, R)|}{p(R) / |J(R)|}.
\end{align}

Taking the ratio of these two conditional densities:
\begin{align}
\frac{\tilde{p}(\tilde{S} \mid \tilde{R} \cup \{\tilde{x}^\star\})}
{\tilde{p}(\tilde{S} \mid \tilde{R})}
\;&=\;
\frac{p(S, R \cup \{x^\star\})}{p(R \cup \{x^\star\})}
\cdot
\frac{p(R)}{p(S, R)}
\nonumber \\
&\quad \cdot
\underbrace{
\frac{|J(R \cup \{x^\star\})| \cdot |J(S, R)|}
{|J(S, R \cup \{x^\star\})| \cdot |J(R)|}
}_{=\,1}.
\end{align}
The Jacobian factors cancel because the joint samples $(S, R \cup \{x^\star\})$ and $(S, R)$ differ by a single point whose Jacobian appears identically in the numerator of one ratio and the denominator of the other. This leaves
\begin{equation}
\frac{\tilde{p}(\tilde{S} \mid \tilde{R} \cup \{\tilde{x}^\star\})}{\tilde{p}(\tilde{S} \mid \tilde{R})}
\;=\; \frac{p(S \mid R \cup \{x^\star\})}{p(S \mid R)}.
\end{equation}
Taking logarithms yields $\mathcal{I}(g(x^\star), g(R), g(S)) = \mathcal{I}(x^\star, R, S)$.
\end{proof}

\subsection{Proof of Theorem~\ref{thm:closedform} (Closed-Form Approximation)}
\label{app:thm1_proof}

\begin{proof}
Let $\hat{p}_R$ denote a Gaussian KDE fit on $R$ with bandwidth $h$, where $|R| = m$. The proof proceeds in four steps.

\textbf{Step 1: Exact decomposition of the augmented KDE.} By the definition of the Gaussian KDE,
\begin{align}
\hat{p}_{R \cup \{x^\star\}}(s)
\;&=\;
\frac{1}{m+1}\sum_{x \in R \cup \{x^\star\}} K_h(s - x)
\nonumber \\
&=\;
\frac{m}{m+1}\,\hat{p}_R(s)
\;+\;
\frac{1}{m+1}\,K_h(s - x^\star).
\end{align}
This identity is exact; no approximation is used.

\textbf{Step 2: Ratio form.} Dividing both sides by $\hat{p}_R(s)$,
\begin{align}
\frac{\hat{p}_{R \cup \{x^\star\}}(s)}{\hat{p}_R(s)}
\;&=\;
\frac{m}{m+1}
+
\frac{1}{m+1}\cdot\frac{K_h(s - x^\star)}{\hat{p}_R(s)}
\nonumber \\
&=\;
1 + \frac{1}{m+1}
\left[
\frac{K_h(s - x^\star)}{\hat{p}_R(s)} - 1
\right].
\end{align}
\textbf{Step 3: Taylor expansion of the logarithm.} Let
\begin{equation}
u_s \;:=\; \frac{1}{m+1}\left[\frac{K_h(s - x^\star)}{\hat{p}_R(s)} - 1\right].
\end{equation}
The regularity assumption $\hat{p}_R(s) \geq c > 0$ on $S_k(x^\star)$, together with the boundedness of the Gaussian kernel $K_h$, implies that $|u_s| \leq C/(m+1)$ for a constant $C$ depending on $c$, $h$, and $d$. For $|u| \to 0$, the Taylor expansion
\begin{equation}
\log(1 + u) \;=\; u - \tfrac{u^2}{2} + O(u^3)
\end{equation}
gives
\begin{equation}
\log\frac{\hat{p}_{R \cup \{x^\star\}}(s)}{\hat{p}_R(s)}
\;=\; \frac{1}{m+1}\left[\frac{K_h(s - x^\star)}{\hat{p}_R(s)} - 1\right] + O(m^{-2}).
\end{equation}

\textbf{Step 4: Summation over the local neighborhood.} Summing over $s \in S_k(x^\star)$:
\begin{align}
\hat{\mathcal{I}}(x^\star; R, S)
\;&=\;
\sum_{s \in S_k(x^\star)}
\log\frac{\hat{p}_{R \cup \{x^\star\}}(s)}{\hat{p}_R(s)}
\nonumber \\
&=\;
\frac{1}{m+1}
\sum_{s \in S_k(x^\star)}
\left[
\frac{K_h(s - x^\star)}{\hat{p}_R(s)} - 1 
\right]+ O(k\,m^{-2})
\nonumber \\
\end{align}
For fixed $k$ or $k = o(m)$, the remainder is $O_p(m^{-2})$, which completes the proof.
\end{proof}

\subsection{Proof of Theorem~\ref{thm:meangap} (Mean Score Gap)}
\label{app:thm2_proof}

\begin{proof}[Proof sketch]
We work throughout with the leading-order expression from Theorem~\ref{thm:closedform}:
\begin{equation}
\hat{\mathcal{I}}(x^\star; R, S) \;\approx\; \frac{1}{m+1} \sum_{s \in S_k(x^\star)}\left[\frac{K_h(s - x^\star)}{\hat{p}_R(s)} - 1\right].
\label{eq:proof2leading}
\end{equation}
The proof proceeds in four steps.

\textbf{Step 1: Expected score.} Let $g(s, x^\star) := K_h(s - x^\star)/\hat{p}_R(s) - 1$. Taking expectations over the randomness in $S$:
\begin{equation}
\mathbb{E}[\hat{\mathcal{I}}(x^\star; R, S)] \;\approx\; \frac{k}{m+1}\cdot \mathbb{E}_{s \sim p_G}\!\left[g(s, x^\star)\,\big|\,s \in S_k(x^\star)\right].
\end{equation}
The $k$-NN localization concentrates $s$ near $x^\star$, where the kernel $K_h(s - x^\star)$ is appreciable; outside this region $g(s, x^\star) \approx -1$ and contributes only to the constant baseline. For the purpose of comparing $\mu_1$ and $\mu_0$, we can treat the expectation as being taken with respect to $p_G$ itself; localization affects $\mu_0$ and $\mu_1$ symmetrically and does not contribute to their difference.

\textbf{Step 2: Substitution of the overfitting decomposition.} Plugging Equation~\ref{eq:overfitmodel} into $\mathbb{E}_{s \sim p_G}[K_h(s - x^\star)/\hat{p}_R(s)]$ yields three additive pieces:
\begin{align}
\mathbb{E}_{s \sim p_G}\!\left[\frac{K_h(s - x^\star)}{\hat{p}_R(s)}\right]
&= \underbrace{\int \frac{K_h(s - x^\star)}{\hat{p}_R(s)}\, p(s)\, ds}_{\text{Piece A: population}} \nonumber \\
&\quad + \underbrace{\frac{\epsilon}{|T|}\sum_{i=1}^{|T|} \int \frac{K_h(s - x^\star)}{\hat{p}_R(s)}\, \phi_\rho(s; x_i)\, ds}_{\text{Piece B: memorization sum}} \nonumber \\
&\quad + \underbrace{\int \frac{K_h(s - x^\star)}{\hat{p}_R(s)}\, r(s)\, ds}_{\text{Piece C: residual}}.
\end{align}

\textbf{Step 3: Differencing member vs.\ non-member expectations.} We examine how each piece changes between the hypotheses $H_0$ ($x^\star \sim P$, independent of $T$) and $H_1$ ($x^\star \in T$).

\emph{Piece A} is a functional of $x^\star$ and the population density $p$ only; it does not reference $T$. Under both hypotheses, $x^\star$ has marginal distribution $P$, so Piece A contributes identically to $\mu_0$ and $\mu_1$ and cancels in the difference. The same argument applies to \emph{Piece C}: the residual $r$ is a property of the generator's output density, not of $x^\star$'s membership status.

For \emph{Piece B}, define the kernel-overlap integral
\begin{equation}
f(x^\star, x_i) \;:=\; \int \frac{K_h(s - x^\star)}{\hat{p}_R(s)}\, \phi_\rho(s; x_i)\, ds,
\end{equation}
which measures the overlap between the kernel centered at $x^\star$ and the memorization bump centered at $x_i$. This integral is largest when $x_i = x^\star$; we denote this \emph{self-overlap} by $\Psi(h, \rho, x^\star) := f(x^\star, x^\star)$. Piece B then reads $\frac{\epsilon}{|T|}\sum_{i=1}^{|T|} f(x^\star, x_i)$, and we compare this sum under the two hypotheses.

Let $B(x^\star) := \sum_{i: x_i \neq x^\star} f(x^\star, x_i)$ denote the \emph{background sum}: the total overlap between the kernel at $x^\star$ and memorization bumps centered at training points \emph{other than} $x^\star$. Then:
\begin{itemize}
    \item \emph{Under $H_0$} ($x^\star \sim P$, independent of $T$): none of the $x_i$ equals $x^\star$, so $\sum_{i=1}^{|T|} f(x^\star, x_i) = B(x^\star)$, a sum of $|T|$ terms.
    \item \emph{Under $H_1$} ($x^\star = x_j$ for some $j$): the sum decomposes as
    \begin{equation}
    \sum_{i=1}^{|T|} f(x^\star, x_i) \;=\; \underbrace{f(x^\star, x_j)}_{=\,\Psi(h, \rho, x^\star)} \;+\; \underbrace{\sum_{i \neq j} f(x^\star, x_i)}_{=\,B(x^\star)},
    \end{equation}
    where the first term is the self-overlap guaranteed by the memorization bump $\phi_\rho(\cdot; x_j)$ sitting on top of the kernel at $x^\star = x_j$, and the second term is a background sum of $|T|-1$ terms.
\end{itemize}
The two versions of $B(x^\star)$ differ by a single term out of $|T|$, which is $O(1/|T|)$ relative to the full background; to leading order they are equal. The dominant contribution to $\mu_1 - \mu_0$ is therefore the self-overlap term, which is present only under $H_1$:

\begin{align}
\mu_1 - \mu_0
\;&\gtrsim\;
\frac{k}{m+1}\cdot\frac{\epsilon}{|T|}
\cdot\Psi(h, \rho, x^\star)
\nonumber \\
&=\;
\frac{k\,\epsilon}{(m+1)\,|T|}
\cdot
\mathbb{E}_s
\!\left[
\frac{
K_h(s - x^\star)\,\phi_\rho(s; x^\star)
}{
\hat{p}_R(s)
}
\right].
\end{align}
\textbf{Step 4: Positivity of the signal term.} The integrand of $\Psi(h, \rho, x^\star)$ is a product of non-negative quantities. It is strictly positive whenever the supports of $\phi_\rho(\cdot; x^\star)$ and $K_h(\cdot - x^\star)$ intersect---i.e., whenever the attacker's bandwidth reaches the memorization region.
\end{proof}

A member $x^\star \in T$ exhibits extra local synthetic density compared to a non-member because the memorization bump $\phi_\rho(\cdot; x^\star)$ is placed at $x^\star$'s own location. All other contributions to local synthetic density, the population term, the residual, and memorization bumps placed at other training points, occur symmetrically for members and non-members (both of which are distributed according to $P$) and cancel in the difference. The attack's power thus derives entirely from the self-overlap signal.

\section{Experiments/ Replication Details}
\label{app:benchmark_details}
\subsection{MIAs for Generative Models Descriptions}
\label{app:mia}

The Membership Inference Attacks referenced in this paper is are described as follows:

\textbf{Distance to Closest Record (DCR / DCR-Diff).}
A common class of membership inference attacks relies on the hypothesis that synthetic generators may reproduce training records or place generated points unusually close to them in feature space \cite{ganleaks}. Under this intuition, query points corresponding to members should exhibit smaller distances to the synthetic dataset than non-members. The Distance to Closest Record (DCR) attack quantifies this effect using the score
\[
f_{\text{DCR}}(x^*,S) = -\min_{x \in S} d(x^*,x),
\]
where \(d(\cdot,\cdot)\) denotes a chosen distance function. A natural extension, DCR-Diff, incorporates a reference dataset \(R\) to normalize this proximity signal by comparing closeness to both datasets:
\[
f_{\text{DCR}}(x^*,S,R) = -\min_{x \in S} d(x^*,x) - \min_{x \in R} d(x^*,x).
\]

\textbf{DOMIAS / Density Estimation.}
DOMIAS \cite{vanbreugel2023membership} approaches membership inference from a distributional perspective by comparing how likely a query record is under the synthetic distribution relative to a reference distribution. Specifically, it evaluates the density ratio
\[
f_{\text{DOMIAS}}(x^*,S,R) = \frac{p_S(x^*)}{p_R(x^*)},
\]
where \(p_S\) and \(p_R\) are density estimates fit to the synthetic and reference datasets, respectively. In practice, these densities may be estimated using approaches such as kernel density estimation or neural density models. A simpler related baseline proposed in \cite{houssiau2022tapas} uses only the synthetic density estimate,
\[
f_{\text{Density Estimate}}(x^*,S) = p_S(x^*),
\]
without explicitly contrasting against a reference distribution.

\textbf{Data Plagiarism Index (DPI) / Local Neighborhood.}
The Data Plagiarism Index (DPI) \cite{ward2024dataplagiarismindexcharacterizing} focuses on localized memorization by examining the composition of the neighborhood surrounding a query point. Given \(x^*\), the method constructs a \(k\)-nearest-neighbor set \(D(x^*)\) drawn jointly from synthetic and reference data, and computes the ratio of synthetic to reference neighbors:
\[
f_{\text{DPI}}(x^*, S, R) =
\frac{\sum_{\mathbf{z} \in D(x^*)}\mathbb{I}(\mathbf{z}\in S)}
{\sum_{\mathbf{z} \in D(x^*)}\mathbb{I}(\mathbf{z}\in R)}.
\]
Higher values indicate stronger local resemblance to the synthetic data distribution. A closely related local attack in \cite{houssiau2022tapas} replaces the fixed-\(k\) neighborhood with all points falling within a specified radius of \(x^*\).

\textbf{LOGAN / Classifier-based.}
LOGAN was first introduced as a white-box membership inference attack \cite{Hayes2017LOGANMI} and later adapted to the black-box setting in \cite{vanbreugel2023membership}. The core idea is to train a classifier to distinguish samples from the target synthetic dataset \(S\) and a reference dataset \(R\). In the original formulation, this classifier is implemented as the discriminator of a GAN trained on synthetic data, and the resulting score
\[
f_{\text{LOGAN}}(x^*,S,R)=D_\theta(x^*)
\]
is used as the membership signal. Records receiving higher discriminator confidence are considered more likely to be members. Subsequent work \cite{houssiau2022tapas} generalized this approach by replacing the GAN discriminator with standard supervised classifiers such as random forests.

\subsection{Generative Model Architecture Descriptions}\label{app:models}
In all experiments, we use the implementations of these models from the Python package Synthcity \cite{synthcity}. For benchmarking purposes, we use the default hyperparameters for each model. A brief description of each model is as follows:
\begin{itemize}
\item \textbf{CTGAN} \cite{Xu2019ModelingTD}: Conditional Tabular Generative Adversarial Network uses a GAN framework with conditional generator and discriminator to capture multi-modal distributions. It employs mode normalization to better learn mixed-type distributions.
\item \textbf{TVAE} \cite{Xu2019ModelingTD}: Tabular Variational Auto-Encoder is similar to CTGAN in its use of mode normalizing techniques, but instead of a GAN architecture, it employs a Variational Autoencoder.

\item \textbf{Normalizing Flows (NFlows)} \cite{durkan2019neural}: Normalizing flows transform a simple base distribution (e.g., Gaussian) into a more complex one matching the data by applying a sequence of invertible, differentiable mappings.

\item \textbf{Bayesian Network (BN)} \cite{Ankan2015}: Bayesian Networks use a Directed Acyclic Graph to represent the joint probability distribution over variables as a product of marginal and conditional distributions. It then samples the empirical distributions estimated from the training dataset.

\item \textbf{Adversarial Random Forests (ARF)} \cite{pmlr-v206-watson23a}: ARFs extend the random forest model by adding an adversarial stage. Random forests generate synthetic samples which are scored against the real data by a discriminator network. This score is used to re-train the forests iteratively.

\item \textbf{Tab-DDPM} \cite{tabddpm}: Tabular Denoising Diffusion Probabilistic Model adapts the DDPM framework for image synthesis. It iteratively refines random noise into synthetic data by learning the data distribution through gradients of a classifier on partially corrupted samples with Gaussian noise.

\item \textbf{PATEGAN} \cite{yoon2018pategan}: The PATEGAN model uses a neural encoder to map discrete tabular data into a continuous latent representation which is sampled from during generation by the GAN discriminator and generator pair.

\item \textbf{Ads-GAN} \cite{yoon2020anonymization}: Ads-GAN uses a GAN architecture for tabular synthesis but also adds an identifiability metric to increase its ability to not mimic training data.
\item \textbf{TabSyn} \cite{tabsyn}: TabSyn uses a Variational Auto-Encoder to learn a latent space in which it builds a diffusion model from. TabSyn usually achieves state of the art data quality metrics relative to other methods compared.
\end{itemize}
\subsection{Benchmarking Datasets References}\label{app:datasets}

\begin{table*}
    \caption{List of Datasets included in Section 5.}
    \centering
    \small
\begin{tabular}{lrrrrr}
\hline
Dataset & OpenML ID & N-size & Classes & Cat. Feat. & Num Feat. \\
\hline
GesturePhaseSegmentationProcessed & 4538 & 9873 & 5 & 1 & 32 \\
MiceProtein & 40966 & 1080 & 8 & 5 & 77 \\
PhishingWebsites & 4534 & 11055 & 2 & 31 & 0 \\
adult & 1590 & 48842 & 2 & 9 & 6 \\
analcatdata\_authorship & 40983 & 4839 & 2 & 1 & 5 \\
bank-marketing & 1461 & 45211 & 2 & 10 & 7 \\
banknote-authentication & 1462 & 1372 & 2 & 1 & 4 \\
blood-transfusion-service-center & 1464 & 748 & 2 & 1 & 4 \\
car & 40975 & 1728 & 4 & 7 & 0 \\
churn & 40701 & 5000 & 2 & 5 & 16 \\
climate-model-simulation-crashes & 1467 & 540 & 2 & 1 & 20 \\
cmc & 23 & 1473 & 3 & 8 & 2 \\
connect-4 & 40668 & 67557 & 3 & 43 & 0 \\
credit-approval & 29 & 690 & 2 & 10 & 6 \\
credit-g & 31 & 1000 & 2 & 14 & 7 \\
diabetes & 37 & 768 & 2 & 1 & 8 \\
electricity & 151 & 45312 & 2 & 2 & 7 \\
eucalyptus & 43924 & 736 & 5 & 15 & 5 \\
kc1 & 1067 & 2109 & 2 & 1 & 21 \\
kc2 & 1063 & 522 & 2 & 1 & 21 \\
kr-vs-kp & 3 & 3196 & 2 & 37 & 0 \\
letter & 6 & 20000 & 26 & 1 & 16 \\
mfeat-morphological & 18 & 2000 & 10 & 1 & 6 \\
numerai28.6 & 23517 & 96320 & 2 & 1 & 21 \\
optdigits & 28 & 5620 & 10 & 1 & 64 \\
pc3 & 1044 & 10936 & 3 & 4 & 24 \\
pendigits & 32 & 10992 & 10 & 1 & 16 \\
phoneme & 1489 & 5404 & 2 & 1 & 5 \\
satimage & 182 & 6430 & 6 & 1 & 36 \\
segment & 40984 & 2310 & 7 & 1 & 19 \\
sick & 38 & 3772 & 2 & 23 & 7 \\
spambase & 44 & 4601 & 2 & 1 & 57 \\
steel-plates-fault & 40983 & 4839 & 2 & 1 & 5 \\
texture & 40499 & 5500 & 11 & 1 & 40 \\
tic-tac-toe & 50 & 958 & 2 & 10 & 0 \\
vehicle & 54 & 846 & 4 & 1 & 18 \\
\hline
\end{tabular}
    \label{tab:datasets}
\end{table*}

\section{Ablations}
\label{app:ablations}
The ablations in for the deep learning and encoding experiments were conducted on a smaller-scale version of the experimental protocol from Section~\ref{sec:experiments}. Across the 15 datasets listed below, each generator was trained on $\min(0.8 \cdot |D|, N)$ records, with synthetic, holdout, and reference sets each drawn at the same size. We use $N = 1000$ for the encoding ablation (Appendix~\ref{app:encoding}) and $N = 250$ for the deep learning ablation (Appendix~\ref{app:dl}) to induce greater overfitting in the latter setting. All other aspects of the protocol, including generator hyperparameters and attack configurations, match the main benchmark. The reduced scale lets us sweep design choices that would be prohibitively expensive at full benchmark size while preserving the qualitative conditions under which Gen-LRA is intended to be deployed.
Datasets:
\begin{enumerate}
    \item \href{https://www.openml.org/search?type=data&sort=runs&id=183&status=active}{Abalone} (OpenML)
    \item \href{https://archive.ics.uci.edu/dataset/2/adult}{Adult}~\cite{adult}
    \item \href{https://archive.ics.uci.edu/dataset/602/dry+bean+dataset}{Bean} (UCI)
    \item \href{https://www.kaggle.com/datasets/shrutimechlearn/churn-modelling}{Churn-Modeling} (Kaggle)
    \item \href{https://archive.ics.uci.edu/dataset/198/steel+plates+faults}{Faults} (UCI)
    \item \href{https://archive.ics.uci.edu/dataset/372/htru2}{HTRU} (UCI)
    \item \href{https://www.kaggle.com/datasets/uciml/indian-liver-patient-records}{Indian Liver Patient} (Kaggle)
    \item \href{https://www.kaggle.com/datasets/mirichoi0218/insurance}{Insurance} (Kaggle)
    \item \href{https://www.kaggle.com/datasets/abhinand05/magic-gamma-telescope-dataset}{Magic} (Kaggle)
    \item \href{https://archive.ics.uci.edu/dataset/332/online+news+popularity}{News} (UCI)
    \item \href{https://www.kaggle.com/datasets/heitornunes/nursery}{Nursery} (Kaggle)
    \item \href{https://www.kaggle.com/datasets/tathagatbanerjee/obesity-dataset-uci-ml}{Obesity} (Kaggle)
    \item \href{https://www.kaggle.com/datasets/henrysue/online-shoppers-intention}{Shoppers} (Kaggle)
    \item \href{https://www.kaggle.com/c/titanic/data}{Titanic} (Kaggle)
    \item \href{https://www.openml.org/search?type=data&sort=runs&id=40983&status=active}{Wilt} (OpenML)
\end{enumerate}

\subsection{Gen-LRA Encoding}
\label{app:encoding}
As our main experiment uses Kernel Density Estimation (KDE) over (usually) heterogeneous datasets, we present an ablation for encoding tabular data to be numeric such that KDE can converge. We experiment with 3 common strategies used in the density estimation literature: ordinal encoding for categorical variables, one-hot encoding categorical variables and then performing Principle Component Analysis (PCA), and using a Variational Auto-Encoder to learn continuous latent representations of the data. 

We repeat our ablation subset experiment with these three encoding schemes. For PCA we use the number of eigenvectors that explain up to 95 \%variance and for the VAE encoding we use TabSyn's original auto-encoder with default settings. Overall, we evaluate the top 100 runs for each metric and find that Ordinal encodings yield the best results (see Table \ref{table:encoding}).

\begin{table*}
\centering
\caption{Mean (std) LRA detection performance by encoding strategy, top 100 runs per metric, averaged over datasets, generative methods, and seeds. Rows sorted by AUC-ROC. Bold marks best encoding per metric.}
\label{table:encoding}
\begin{tabular}{cccccc}
\toprule
Encoding & AUC-ROC & TPR@FPR=0 & TPR@FPR=0.001 & TPR@FPR=0.01 & TPR@FPR=0.1 \\
\midrule
Ordinal & \textbf{0.616 (0.053)} & \textbf{0.011 (0.010)} & \textbf{0.013 (0.012)} & \textbf{0.043 (0.028)} & \textbf{0.221 (0.079)} \\
VAE & 0.594 (0.049) & 0.009 (0.010) & 0.012 (0.012) & 0.039 (0.027) & 0.198 (0.064) \\
PCA & 0.575 (0.046) & 0.010 (0.008) & 0.012 (0.009) & 0.031 (0.021) & 0.171 (0.056) \\
\bottomrule
\end{tabular}
\end{table*}
\subsection{Deep Learning Estimation}
\label{app:dl}
We replace the KDE surrogate in Gen-LRA with Block Neural Autoregressive Flows (BNAF)~\citep{bnaf19}, following the implementation and hyperparameters of~\cite{vanbreugel2023membership}. Each BNAF model is trained with default implementation parameters with early stopping on a held-out validation split of the reference set, using the Adam optimizer with default learning rate. We fit one BNAF on $R$ and a second on $R \cup \{x^\star\}$ for each test point and compute the Gen-LRA score using these two surrogates in place of the corresponding KDEs. To induce greater overfitting in the trained generators and thereby produce stronger membership signal for both surrogates to recover, we reduce the training, holdout, reference, and synthetic set sizes to 250 records each. We then evaluate the top 100 highest-scored runs for each metric from the ordinal-encoded KDE and BNAF versions of Gen-LRA. All other aspects of the small-scale benchmark protocol, including datasets, generators, and attack baselines, match Section~\ref{app:ablations}. Results and analysis appear in Section~\ref{subsec:deeplearningdensityestimation}.

\subsection{Ablation: Different $k$ sizes}\label{app:k}
Gen-LRA targets local overfitting by utilizing the \( k \)-nearest neighbors in \( S \) to \( x^* \). Consequently, \( k \) serves as a hyperparameter in the attack. To assess the impact of \( k \) on attack efficacy, we replicate the benchmarking experiments from Section \ref{sec:experiments} across varying values of \( k \). The average AUC-ROC and corresponding standard deviations for the top 300 runs for each evaluation metric are reported in Table \ref{table:k_values}. Empirically, we observe that smaller values of \( k \) generally enhance attack performance, though this effect varies by model.

\begin{table*}
\small
\caption{Mean AUC-ROC at different $k$ values for Gen-LRA for top 300 highest runs.} 
\centering 

\begin{tabular}{lllllll}
\toprule
Model & $k=1$ & $k=5$ & $k=10$ & $k=25$ & $k=50$ &  $k=100$ \\
\midrule
AdsGAN & 0.57 (0.00) & 0.58 & 0.57 (0.00) & 0.57 (0.01) & 0.58 (0.01) & 0.61 (0.00)\\
ARF & 0.60 (0.02) & 0.58 (0.02) & 0.59 (0.01) & 0.58 (0.01) & 0.58 (0.01) & 0.58 (0.01) \\
CTGAN & 0.60 (0.01) & 0.58 (0.00) & 0.59 (0.02) & 0.60 (0.00) & 0.58 (0.01) & 0.58 (0.01) \\
Tab-DDPM & 0.60 (0.02) & 0.60 (0.01) & 0.60 (0.02) & 0.61 (0.00) & 0.60 (0.01) & 0.60 (0.01) \\
N-Flow & 0.61 (0.00) & 0.57 (0.01) & 0.59 (0.00)& 0.59 (0.00)& 0.59 (0.00) & 0.60 (0.01) \\
RTF & 0.59 (0.02) & 0.59 (0.02) & 0.59 (0.02) & 0.60 (0.02) & 0.59 (0.02) & 0.59 (0.02) \\
TabSyn & 0.62 (0.03) & 0.61 (0.02) & 0.60 (0.02) & 0.60 (0.01) & 0.59 (0.01) & 0.58 (0.01) \\
TVAE & 0.60 (0.00) & 0.59 (0.00) & 0.57 (0.00) & 0.57 (0.00) & 0.58 (0.01) & 0.58 (0.01) \\
\bottomrule
\end{tabular}
\label{table:k_values}
\end{table*}

\section{Additional Figures}
\begin{table*}[t]
\caption{Median with interquartile ranges for MIAs across the top 100 runs for each metric. Distance-based baselines (DCR, DCR-Diff) exhibit medians well below their means, with IQRs touching zero, indicating that their strong-run performance is driven by a small set of near-exact memorization outliers. Gen-LRA's median exceeds its own mean and its IQR is bounded away from zero, with median AUC (0.594) exceeding the third quartile of almost every baseline.}\centering
\small
\setlength{\tabcolsep}{4pt}
\begin{tabular}{lccccc}
\toprule
& \multirow{2}{*}{} & \multicolumn{4}{c}{TPR at Fixed FPR} \\
\cmidrule(lr){3-6}
Method & AUC & 0.0 & 0.001 & 0.01 & 0.1 \\
\midrule
Gen-LRA & \textbf{0.594 [0.574, 0.609]} & \textbf{0.055 [0.034, 0.067]} & \textbf{0.055 [0.034, 0.067]} & \textbf{0.055 [0.038, 0.072]} & \textbf{0.214 [0.175, 0.237]} \\
Classifier & 0.559 [0.524, 0.597] & 0.005 [0.000, 0.017] & 0.005 [0.001, 0.018] & 0.015 [0.007, 0.027] & 0.133 [0.105, 0.181] \\
DCR & 0.579 [0.561, 0.600] & 0.011 [0.000, 0.053] & 0.012 [0.002, 0.053] & 0.022 [0.011, 0.063] & 0.178 [0.125, 0.212] \\
DCR-Diff & 0.576 [0.541, 0.591] & 0.024 [0.008, 0.063] & 0.025 [0.008, 0.063] & 0.031 [0.013, 0.072] & 0.169 [0.131, 0.205] \\
DOMIAS & 0.546 [0.523, 0.568] & 0.019 [0.009, 0.034] & 0.019 [0.009, 0.035] & 0.020 [0.008, 0.032] & 0.132 [0.110, 0.172] \\
DPI & 0.528 [0.511, 0.543] & 0.000 [0.000, 0.000] & 0.001 [0.001, 0.001] & 0.011 [0.010, 0.013] & 0.113 [0.101, 0.123] \\
LOGAN & 0.495 [0.474, 0.507] & 0.007 [0.002, 0.020] & 0.006 [0.002, 0.020] & 0.010 [0.002, 0.026] & 0.107 [0.082, 0.123] \\
Local Neighborhood & 0.534 [0.505, 0.557] & 0.005 [0.000, 0.012] & 0.006 [0.001, 0.015] & 0.014 [0.010, 0.029] & 0.118 [0.102, 0.145] \\
MC & 0.553 [0.522, 0.571] & 0.006 [0.000, 0.023] & 0.006 [0.002, 0.024] & 0.017 [0.009, 0.031] & 0.121 [0.100, 0.165] \\
\bottomrule
\end{tabular}

\label{tab:median_iqr}
\end{table*}
\end{document}